\documentclass[10pt,journal,compsoc]{IEEEtran}
\pdfoutput=1

\usepackage{ifpdf}

\ifCLASSOPTIONcompsoc
  \usepackage[nocompress]{cite}
\else
  \usepackage{cite}
\fi
\ifCLASSINFOpdf
   \usepackage[pdftex]{graphicx}
\else
\fi
\usepackage{amsmath}
\usepackage{amssymb}
\usepackage{algorithmic}
\usepackage{algorithm}
\makeatletter
\newcommand{\removelatexerror}{\let\@latex@error\@gobble}
\makeatother

\usepackage{array}
\usepackage{color}

\ifCLASSOPTIONcompsoc
 \usepackage[caption=false,font=footnotesize,labelfont=sf,textfont=sf]{subfig}
\else
 \usepackage[caption=false,font=footnotesize]{subfig}
\fi
\usepackage{url}
\usepackage{multirow}

\usepackage[T1]{fontenc}

\begin{document}
%
\title{Computational Optics for\\ Mobile Terminals in Mass Production}
%
%
%
%

\author{Shiqi~Chen,
        Ting~Lin, 
        Huajun~Feng, 
        Zhihai~Xu,
        Qi~Li,
        and~Yueting~Chen 
\IEEEcompsocitemizethanks{\IEEEcompsocthanksitem Shiqi Chen, Ting Lin, Huajun Feng, Zhihai Xu, Qi Li, and Yueting Chen are with the State Key Laboratory of Modern Optical Instrumentation, Zhejiang University, Hangzhou, 310000, China. 
\IEEEcompsocthanksitem Corresponding author: Yueting Chen (E-mail: chenyt@zju.edu.cn)
\IEEEcompsocthanksitem This work is supported by the Civil Aerospace Pre-Research Project (number D040104) and
the National Natural Science Foundation of China (number 61975175). 
}
\thanks{Manuscript received April 8, 2022.}}

%
%

\markboth{IEEE TRANSACTIONS ON PATTERN ANALYSIS AND MACHINE INTELLIGENCE}%
{Shell \MakeLowercase{\textit{et al.}}: Bare Demo of IEEEtran.cls for Computer Society Journals}
%



\IEEEtitleabstractindextext{%
\begin{abstract}
Correcting the optical aberrations and the manufacturing deviations of cameras is a challenging task. Due to the limitation on volume and the demand for mass production, existing mobile terminals cannot rectify optical degradation. In this work, we systematically construct the perturbed lens system model to illustrate the relationship between the deviated system parameters and the spatial frequency response (SFR) measured from photographs. To further address this issue, an optimization framework is proposed based on this model to build proxy cameras from the machining samples' SFRs. Engaging with the proxy cameras, we synthetic data pairs, which encode the optical aberrations and the random manufacturing biases, for training the learning-based algorithms. In correcting aberration, although promising results have been shown recently with convolutional neural networks, they are hard to generalize to stochastic machining biases. Therefore, we propose a dilated Omni-dimensional dynamic convolution (DOConv) and implement it in post-processing to account for the manufacturing degradation. Extensive experiments which evaluate multiple samples of two representative devices demonstrate that the proposed optimization framework accurately constructs the proxy camera. And the dynamic processing model is well-adapted to manufacturing deviations of different cameras, realizing perfect computational photography. The evaluation shows that the proposed method bridges the gap between optical design, system machining, and post-processing pipeline, shedding light on the joint of image signal reception (lens and sensor) and image signal processing (ISP).
\end{abstract}

}

\maketitle

\IEEEdisplaynontitleabstractindextext

%
\IEEEpeerreviewmaketitle

\IEEEraisesectionheading{\section{Introduction}\label{sec:introduction}}

\begin{figure*}[htbp]
    \centering
    \includegraphics[width=\linewidth]{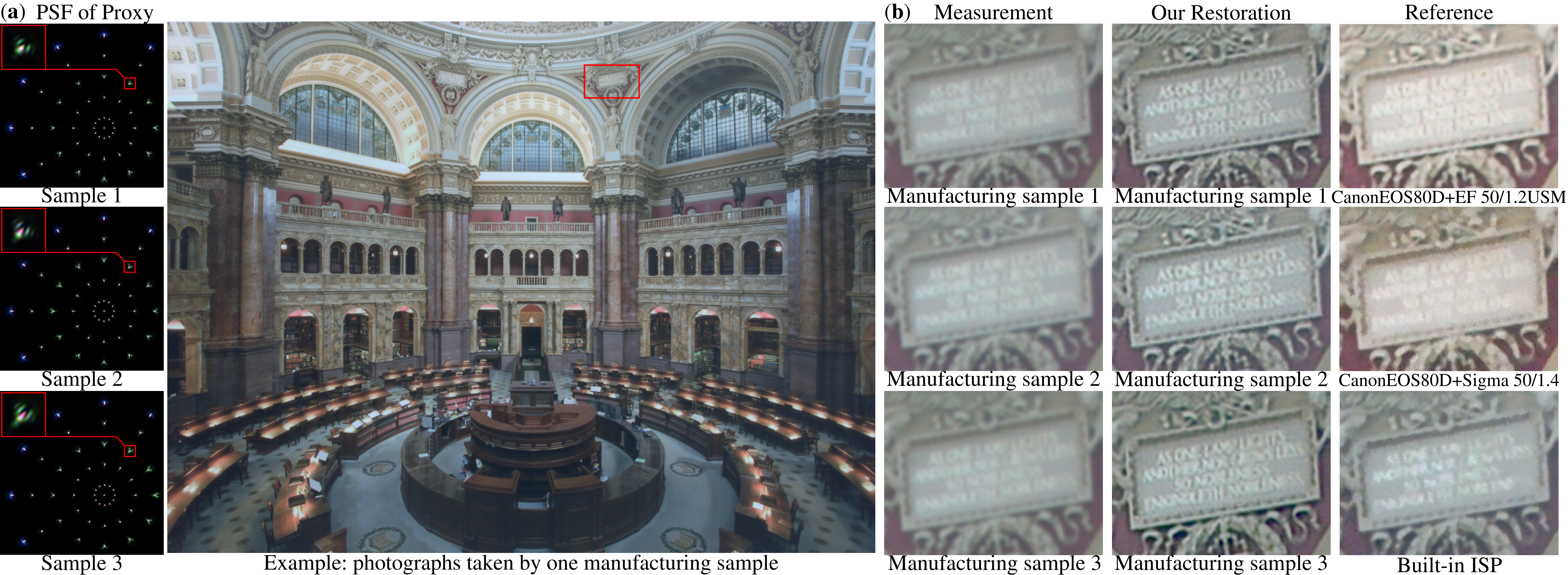}
    \caption{\textbf{Manufacturing biases adaptation and comparisons}. (a) centrosymmetric PSF calculated by the proxy camera of different machining samples \textbf{(Best viewed with zoom)}. (b) magnified comparisons of the photographs taken in the same scene. We show the measured degradation of each \textbf{Phone} (converted to sRGB for visualization) and our restoration output from the same ISP pipeline. And the results of high-end DSLR cameras are shown for reference (captured under same aperture for comparisons).}
    \label{fig:teaser}
\end{figure*}

%
%
%
%
\IEEEPARstart{I}{n} the machining and assembly procedure of imaging systems, deflection and manufacturing bias affect the shape and positions of lenses \cite{lin_novel_2011}. Even subtle shape or position variations will introduce additional aberrations, which significantly degrade the optical performance of cameras \cite{Hsueh:10}. To be more specific, the deflection will lead to the point spread function (PSF) difference (\textbf{Fig.} \ref{fig:teaser}a) of the symmetrical field-of-view (FoV), and the manufacturing deviation will cause the overall decrease in SFR (\textbf{Fig.} \ref{fig:overview}c). Hence, analyzing the biases between the ideal and manufacturing is a critical issue in the optomechanical design of the imaging system, and it is essential for improving processing quality and controlling the cost \cite{Hu:15, Wang:13}.

We hope to estimate the gap between the ideal design and the produced devices, aiming for performing a targeted restoration within ISP systems \cite{Jung:11}. This line of research has promoted significant processes recently \cite{malacara_optical_2007}. Commercially, existing optical design programs have integrated with tolerance analysis to assess the performance of perturbed systems \cite{rimmer_analysis_1970} or calculate the bias range of each parameter according to the measured indicator \cite{rimmer_tolerancing_1978}, \textit{e.g.}, modulation transfer function (MTF). However, these tolerancing procedures still face a few challenges for application in a specific machining sample \cite{lin_novel_2011}. One issue is the tolerance, which is generally selected empirically or according to the performance requirements without considering the actual machining procedure \cite{Sun:14}. Another challenge is the inherent gap between the tolerancing indicator and the measurement of SFR \cite{derby1999optomechanical}. This theoretical difference leads to severer biases predicted by tolerancing programs. In academia, representative works include local optimization \cite{hutchison_image_2012}, which modifies the system parameters by the pixel-level difference of PSF. As well as the end-to-end optical system optimization proposed recently \cite{sun_end--end_2021, Tseng2021DeepCompoundOptics}, which optimizes the system with image-to-image rendering or deep learning model. Nevertheless, the existing methods still suffer from several limitations. For example, these optimizations are significantly affected by the noise in actual measurement \cite{mosleh_camera_2015}. And the differentiable framework requires a large volume of paired data where the ideal image or the optical parameters and the corresponding PSFs are tremendously complicated to acquire \cite{li_end--end_2021}.

This paper is devoted to a fundamental solution to ISP systems - bridging the gap between optics and postprocessing. We show an illustrative example in \textbf{Fig.} \ref{fig:teaser}b, where the measurements of manufacturing samples are slightly different, yet the realized restorations are similar after processing. The classical ISP system is a step-by-step process where each module cascades with each other \cite{nishimura_automatic_2018}. This separation mitigates the processing difficulty of each module but allows the slight errors accumulated in the subsequent operations \cite{heide_flexisp_2014}. To this end, recent works implement end-to-end methods for the mapping from Bayer pattern data to sRGB images \cite{ignatov_replacing_2020}. However, the acquisition of densely-labeled data specified for each camera is time-consuming, and the enormous computational overhead is the critical limitation of deploying it into mobile terminals \cite{ignatov_ntire_2019}.

In this work, we mainly focus on connecting the optical design, the system manufacturing, and the ISP systems. A perturbation model is proposed for describing the influence of the deviated parameters on the geometric image evaluation, \textit{e.g.}, SFR. Distinct from the indicator calculation through exit pupil wavefront (used in optical design programs), we adopt the imaging simulation to obtain realistic photographs and apply SFR to measure MTF, which follows the process of image formation and index evaluation. It also provides an optimization framework to estimate the perturbed parameters from the measured indicators. In this way, a proxy camera, whose imaging results are close to reality, is constructed. It acts as a bridge for the co-design of the actual camera and the subsequent ISP.

Furthermore, we propose a brand new dynamic postprocessing architecture based on DOConv, expecting to handle the degradation of various system perturbations with one model. Engaging with the imaging simulation of the proxy camera, we encode the accumulated errors of cascaded modules and the degradation of system perturbations into the data pairs for the training of the dynamic model. This model realizes the correction of various machining samples' optical degradations with less computational overhead. Our approach can be easily adapted into new camera devices at a penurious cost on indicator measurement, thus bypassing the time-consuming paired data collection.

We evaluate the proposed method on two imaging systems: customized digital single-lens reflex (DSLR) cameras and Huawei Honor 20 Pro (Phone), with known ideal optical parameters of both cameras. The assessments include the accuracy of constructing the proxy camera and the perturbation adaptability of the dynamic postprocessing model as well as the benefits to downstream vision applications (\textit{i.e.}, object detection, optical character recognition). Extensive experiments demonstrate that our method has the potential to link optical machining with postprocessing for realizing targeted restorations. The results are on par or sometimes even outperform the high-end DSLR lenses (the restoration results of Phones are shown in \textbf{Fig.} \ref{fig:teaser}).


Our main contributions are summarized as follows:
\begin{itemize}[\IEEEsetlabelwidth{Z}]
  \item We construct a perturbed optical system model based on the image formation process, which illustrates the relationship between the perturbation of system parameters and the measured SFR. We demonstrate the advantage of the proposed perturbing model over the existing optical tolerancing procedure.

  \item We propose an optimization method to infer the system perturbation from the SFRs measured from actual manufacturing samples, hoping to construct proxy cameras whose imaging results are close to reality. These proxy cameras are to generate the data pairs that characterize the mapping of optical degradation, thereby fast adapting to the data acquisition of new devices in mass production.
  
  \item We propose a dilated Omni-dimensional dynamic convolution (DOConv) and implement it into post-processing to tackle spatially varying aberrations and stochastic machining deviations. It can be embedded into the existing ISP systems and correct the errors accumulated by modules cascade.
\end{itemize}

This paper proceeds as follows. In Section. \ref{sec:related work}, we review the related works. Section. \ref{sec:3} presents the optical perturbation model and the optimization to construct the proxy camera. The dynamic postprocessing pipeline is detailed in Section. \ref{sec:4}. Quantitative and qualitative experimental analyses of our approach are provided in Section. \ref{sec:5}. Section. \ref{sec:6} explores the potential applications of the proposed method. Conclusions and discussions are drawn in Section. \ref{sec:7}.

\section{Related Work}\label{sec:related work}
Optical degradation correction is a comprehensive mission in computational photography, where algorithms and optical systems cooperate. The additional machining bias poses a significant challenge to solving this problem. In this section, we present an overview of end-to-end optical system optimization and ISP systems.

\subsection{End-to-End Optical System Optimization}\label{sec:related area1}
Optical designers generally select the tolerance empirically or according to the performance requirements where the system bias is randomly sampled within tolerance to determine statistical degradation in mass production \cite{Foote:48}. However, this process does not consider the actual machining procedure, which is meaningless for a particular manufacturing sample \cite{lin_novel_2011}. Besides the top-down tolerance analysis, considerable works infer the system parameters by polynomial (Zernike) fitting or convex optimization \cite{rimmer_analysis_1970, rimmer_tolerancing_1978, Tseng2021NeuralNanoOptics}. For polynomial fitting, the predicted range is generally broader than the actual for the multiple coefficients system \cite{10.1117/12.55675}. Some works have taken advantage of the measured PSFs or images to fine-tune the entire system and perform targeted restoration \cite{hutchison_image_2012}. However, the noise introduced in actual measurement can easily affect them since these methods are guided by the pixel-by-pixel mean squared errors (MSE) evaluation \cite{mosleh_camera_2015}. Recently, some works proposed to jointly optimize the optical parameters and the postprocessing systems in a differentiable manner, where the model of Fourier Optics \cite{9466261}, the image-to-image rendering \cite{sun_end--end_2021}, the proxy deep-learning model \cite{Tseng2021DeepCompoundOptics} are used to build the end-to-end pipeline. However, the end-to-end optimizations have some limitations when applied to complex systems with large FoV  (in-depth discussions in the \textbf{supplementary file}).

We modify the ideal system to construct a proxy camera, whose system bias may not be precise compared to the actual device, but their imaging results are close. The proposed method guides the optimization by the geometric optical image evaluation, which is less susceptible to noise. We demonstrate that our physical-based framework can generate realistic imaging results for various machining samples and work under different noise levels.

\subsection{Mobile ISP Systems}\label{sec:related area2}
Considerable efforts have been invested in image postprocessing to correct optical degradation \cite{chen2018learning, zhang2019zoom, hasinoff2016burst, yan2019attention, 10.1145/3474088, Chen_2021_ICCV}. Traditional deconvolution approaches utilize multiple image priors for iterative or mutual optimization to obtain the latent images \cite{heide2013high, pan2016blind}. Unfortunately, they are inefficient in dealing with spatially varying degradation and thus have difficulty applying to real-time imaging \cite{sun2017revisiting}. The existing mobile ISP system divides the processing into multiple steps: white balance \cite{van2007edge, gijsenij2011improving, afifi_deep_2020}, denoising \cite{buades2005non, condat2010simple}, Bayer pattern interpolation \cite{hirakawa2005adaptive, dubois2006filter, li2008image}, color correction \cite{kwok2013simultaneous}, \textit{etc}. Separating the task into independent modules facilitates the processing overhead, but the error of one module will be accumulated and magnified in subsequent steps, resulting in the wrong outputs \cite{ignatov_replacing_2020}.

Recent works propose to replace the cascaded ISP systems with deep learning models to address this issue \cite{mei_higher-resolution_2019, ignatov_ntire_2019}. Such models are entirely data-driven and also have the potential for real-time imaging. \cite{ignatov_replacing_2020} proposes to collect the data pairs by shooting the same scene with a mobile phone and a high-end DSLR. However, this data construction is time-consuming and has poor portability for new devices \cite{zhang_star_2021}. Recent works exploit to obtain data pairs by imaging simulation of an ideal system \cite{10.1145/3474088}, but they do not consider the machining bias introduced during manufacturing. Therefore, there is a particular domain gap between the training data and the real-shot images, resulting in unsatisfied generalization for any processing samples in the actual scene \cite{liang_cameranet_2019}.

Machining degradations introduced during camera production increase the difficulty of postprocessing algorithms. The deep learning methods mentioned above are fixed in the inference and cannot adaptively deal with the degradation of input features \cite{yang_condconv_2020}. We noticed that many approaches apply attention or transformer to endow the network with dynamic processing ability \cite{zhang_dynet_2020, li2021omni, liu_swin_2021, liu_convnet_2022}. On a large scale of data, these models have a better performance than the static model, yet in the data with a relatively single distribution, they tend to overfit \cite{liang_swinir_2021, zamir_restormer_2021}.

In this work, multiple proxy cameras are engaged to synthesize realistic data for the training of the dynamic model. The proposed method successfully restores the optical degradation of complex distribution and realizes adaptive postprocessing of samples with different deviations.

\section{Perturbed Optical System Model}\label{sec:3}
\begin{figure*}[htbp]
    \centering
    \includegraphics[width=\linewidth]{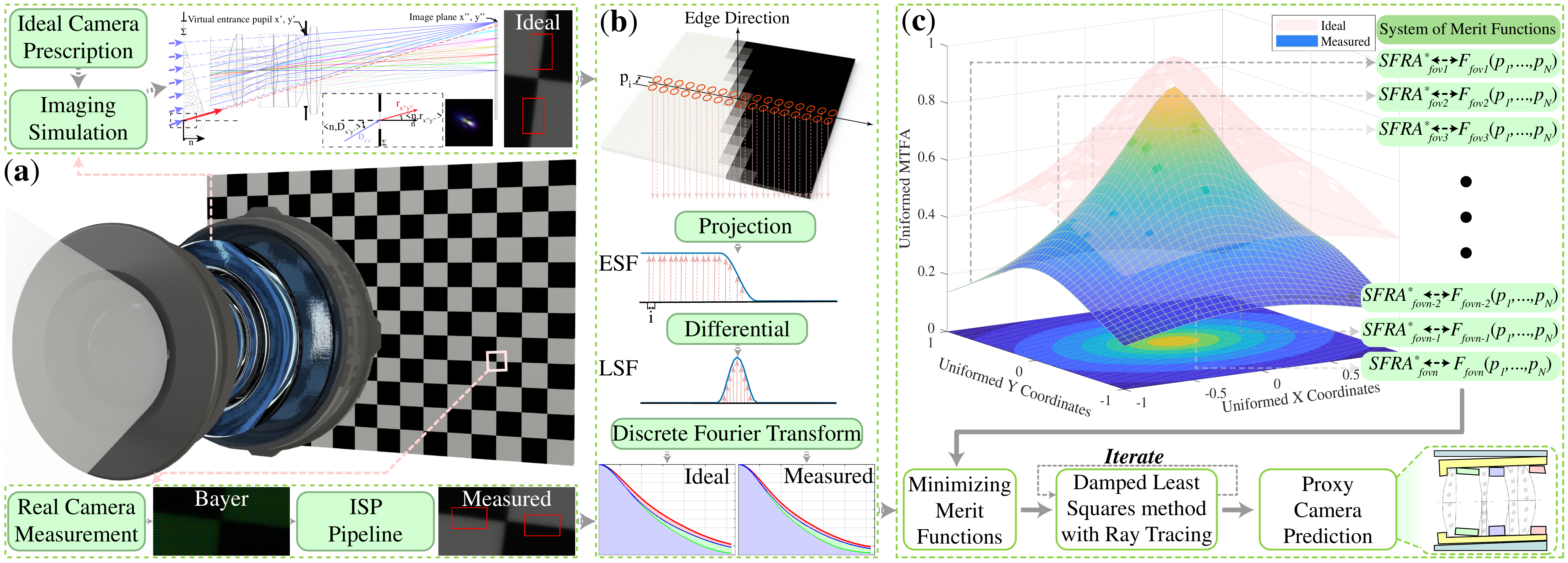}
    \caption{\textbf{Overview of the perturbed optical system model}. (a) We simulate the imaging results of the ideal edge by the camera's parameters (top) and acquire the measured edge by photographing with real devices (bottom). (b) The procedure from edge profile to SFR goes through projection, differential, and DFT. In this way, we obtain the SFRA of ideal design and the measured edge (detailed in Sec. \ref{sec:3.1.2}). (c) Set the measured SFRAs as targets to optimize the system parameters and predict the proxy camera by damped least-squares iteration (detailed in Sec. \ref{sec:3.2}).}
    \label{fig:overview}
\end{figure*}
With the lens prescriptions and the actual manufacturing sample, our goal is to construct a proxy camera whose imaging results are relatively close to the manufactured device. Different from the ideal designing procedure, geometric optical image evaluation such as the SFR generally suffers from the deviations introduced in manufacturing and mounting assembly. Moreover, the discrete sampling and noise of the sensor are non-negligible in measurement. Therefore, in the following Section. \ref{sec:3.1}, we first analyze the image formation procedure of the perturbed optical system, aiming for constructing the relationship between the perturbation of system parameters and the measured SFR. Then in the Section. \ref{sec:3.2}, the method to build a proxy camera is presented in a detailed account.

\subsection{Geometric Optical Image Evaluation}\label{sec:3.1}
A general camera is primarily divided into the optical lens and the photosensor, where the former gathers the scene information and the latter records the intensity of the signal. While due to the inevitable manufacturing deviation, the scene rays collected by the camera will be deflected unexpectedly during propagation, resulting in degraded images and unfavorable indicators. To model the perturbations of the system as well as their influence on the SFR measurement, we consider the case where an incident ray is traced in a camera modeled with biased coefficients. 

\subsubsection{Perturbed System Coefficients}\label{sec:3.1.1}
\noindent \textbf{Ray-surface intersection} The spatial coordinates $(x, y, z)$ of an incident ray are as follows:
\begin{equation}
    \label{eq:incident ray equation}
    x = x_{0}+ks, y = y_{0}+ls, z = z_{0}+ms,
\end{equation}
here $s$ is the parameter of distance along the ray measured from the source point $(x_{0}, y_{0}, z_{0})$, and $D=(k,l,m)$ are the normalized direction vector. The general surface encountered in mobile camera may be represented by:
\begin{equation}
    \label{eq:equation of aspheric}
    \mathit{F}(x, y, z) = z - \frac{\widetilde{c}\rho^{2}}{1 + \sqrt{1-(1+\widetilde{k})\widetilde{c}^{2}\rho^{2}}} - \sum_{j=1}^{\mathcal{N}}\widetilde{A}_{2j}\rho^{2j},
\end{equation}
where $z$ is the coordinate along the optical axis. $\rho=\sqrt{x^{2} + y^{2}}$ is the distance from a surface point to the optical axis. $\widetilde{c}$ is the perturbed vertex curvature and $\widetilde{k}$ is the perturbed conic constant. $\widetilde{A}_{2j}$ is the $2j^{th}$ perturbed power aspherical coefficient where $\mathcal{N}$ is the total order of aspheric. To determine the ray-surface intersection, we apply the Newton-Raphson iteration method to find a value $s$ such that the coordinates value $(x, y, z)$ from Eq. \eqref{eq:incident ray equation} satisfy the surface Eq. \eqref{eq:equation of aspheric}. In every iteration $i$, the distance parameter $s$ is updated by:
\begin{equation}
    \label{eq:newton-raphson iterate}
    s_{i+1} = s_{i} - \mathit{F}(x_{i}, y_{i}, z_{i})/\mathit{F}'(x_{i}, y_{i}, z_{i}),
\end{equation}
where $(x_{i}, y_{i}, z_{i}) = (x, y, z)|_{s_{i}}$ and:
\begin{equation}
    \label{eq:direction partial}
    \mathit{F}'(X_{i}, Y_{i}, Z_{i}) = (F_{x})_{i}k + (F_{y})_{i}l + (F_{z})_{i}m,
\end{equation}
here $(F_{x})_{i}$ denotes $\partial\mathit{F}/\partial x$ evaluated at $(x_{i}, y_{i}, z_{i})$. Similar calculations are performed with respect to $(F_{y})_{i}$ and $(F_{z})_{i}$. The iteration process is terminated with the value $s_{i}$ when
\begin{equation}
    \label{eq:iteration terminate}
    |s_{i} - s_{i-1}| < \epsilon',
\end{equation}
where $\epsilon'$ is a small preassigned value that can be adjusted according to the required accuracy. After the intersection is determined, we follow the Snell's law to carry out the direction vector after refraction:
\begin{equation}
    \label{eq:snell's law}
    \widetilde{n'}\cdot \mathbf{D}'\times \mathbf{r} = \widetilde{n}\cdot\mathbf{D}\times\mathbf{r},
\end{equation}
here $\widetilde{n}$ and $\widetilde{n}'$ are the perturbed refractive indices of the medium where the ray is incident and refracted, which is modeled by the material perturbation of d-light (\textit{i.e.}, refractive index $\widetilde{n}_{d}$ and abbe number $\widetilde{v}_{d}$). For the detailed material perturbation model, please refer to the \textbf{supplementary file}. $\mathbf{D}'=(k', l', m')$ is the unit vectors denoting the direction of refracted ray. $\mathbf{r}=(K, L, M)$ is a normal vector of the surface at the intersection. Indicating by Eq. \eqref{eq:snell's law}, the coplanarity of vector $\mathbf{D}$, $\mathbf{D}'$, and $\mathbf{r}$ allows us to represent $\mathbf{D}'$ by the linear combination of $\mathbf{D}$ and $\mathbf{r}$:
\begin{equation}
    \label{eq:refracted direction}
    \mathbf{D}' = \mu\mathbf{D} + \Gamma\mathbf{r},
\end{equation}
where $\mu=n/n'$ and $\Gamma$ is an undetermined multiplier. Squaring and adding the component of Eq. \eqref{eq:refracted direction}, we obtain a quadratic in $\Gamma$, whose analytical solution is easy to solve:
\begin{equation}
    \label{eq:quadratic in gamma}
    \begin{split}
        & \qquad \qquad \quad \Gamma^{2} + 2a\Gamma + b = 0, \\
        a &= \mu(kK+lL+mM)/(K^{2}+L^{2}+M^{2}), \\
        b &= (\mu^{2}-1)/(K^{2}+L^{2}+M^{2}),
    \end{split}
\end{equation}

\noindent \textbf{Surface-to-surface transfer} Apart from the tracing procedure in a rotationally symmetric system, the tilts of optical elements are significant factors to be reckoned with when light propagates in a real camera. We model the incline of a surface in terms of Euler angles, where three successive rotations $\widetilde{t}_{zx}$, $\widetilde{t}_{yz}$, $\widetilde{t}_{xy}$ are to switch the ray data between the system of the optical axis and element:
\begin{equation}
    \label{eq:system transformation}
    \left(
    \begin{matrix}
        \overline{x} & \overline{k} \\
        \overline{y} & \overline{l} \\
        \overline{z} & \overline{m}
    \end{matrix}
    \right) = R
    \left(
    \begin{matrix}
        x-x_{\xi} & k\\
        y-y_{\xi} & l\\
        z-z_{\xi} & m
    \end{matrix}
    \right),
\end{equation}
here $(x_{\xi}, y_{\xi}, z_{\xi})$ is the origin of the reference system. And the letters denoted by $\overline{(\cdot)}$ are the ray data of the transforming system. $R$ is expressed as follows:
\begin{equation}
    \label{eq:rotation}
    \footnotesize
    \left(
    \begin{matrix}
        \cos{\widetilde{\gamma}} & -\sin{\widetilde{\gamma}} & 0 \\
        \sin{\widetilde{\gamma}} &  \cos{\widetilde{\gamma}} & 0 \\
                   0 &             0 & 1 \\
    \end{matrix}
    \right)
    \left(
    \begin{matrix}
                   1 &             0 &            0 \\
                   0 &   \cos{\widetilde{\beta}} & -\sin{\widetilde{\beta}} \\
                   0 &   \sin{\widetilde{\beta}} &  \cos{\widetilde{\beta}} \\
    \end{matrix}
    \right)
    \left(
    \begin{matrix}
        \cos{\widetilde{\alpha}} &             0 & -\sin{\widetilde{\alpha}} \\
                   0 &             1 &            0 \\
        \sin{\widetilde{\alpha}} &             0 & \cos{\widetilde{\alpha}} \\
    \end{matrix}
    \right),
\end{equation}
where $\widetilde{\alpha}=\widetilde{t}_{zx}$, $\widetilde{\beta}=\widetilde{t}_{yz}$, $\widetilde{\gamma}=\widetilde{t}_{xy}$ are the angles between the reference system and the transforming system. 

After system transformation, the propagation from one surface to the next surface is first addressed by tracing the ray following the perturbed thickness vector $\widetilde{d}$ (represented by the norm $\widetilde{l}$ and the direction cosine $(\widetilde{d}_{x}, \widetilde{d}_{y}, \widetilde{d}_{z})$):
\begin{equation}
    \label{eq:surface-surface propagation}
    s = (\widetilde{l} \cdot \widetilde{d}_{z} - \overline{z})/m, x = \overline{x} + ks - \widetilde{l} \cdot \widetilde{d}_{x}, y = \overline{y} + ls - \widetilde{l} \cdot \widetilde{d}_{y},
\end{equation}

In this way, we model the perturbation of thickness and decenter in the meanwhile. For the detailed decenter illustration, please refer to the \textbf{supplementary file}. After the propagation between two surfaces, the ray data is transformed into the system of the latter surface by Eq. \eqref{eq:system transformation}. And the aforementioned steps are repeated for succeeding surfaces in sequence until the sensor plane, where the ray data and the optical path length of each ray are recorded.

In summary, the perturbed optical system as well as its impact on incident ray are modeled through surface-by-surface ray tracing. We consider the potential perturbation of all optical elements to construct an authentic model based on physical procedure (the specific configurations are listed in the \textbf{supplementary file}). The recorded ray data is used to establish the link between perturbed coefficients and the SFR measurement, which will be illustrated in the following.

\subsubsection{SFR Measurement via Imaging Simulation}\label{sec:3.1.2}
In the optical design process, the MTF of the imaging system is obtained by Fourier transforming the continuous wavefront on the exit pupil plane. However, in the geometric optical image evaluation, the SFR (equivalent to the system's MTF) is generally obtained by the Fourier transform of the line spread function (LSF) captured by the sensor. As demonstrated in the Appendix A of the \textbf{supplementary file}, the discrete sampling of the sensor and the noise will make differences in SFR. Therefore, we adopt the imaging simulation technique to synthesize a realistic edge image to ensure the procedure of calculating SFR between simulation and measurement is similar. In imaging simulation, the diffraction effect caused by the optical aperture is another content to be considered in addition to the aberrations calculated by tracing. So we inverse the tracing ray (calculated in Section. \ref{sec:3.1.1}) from the sensor plane to the exit pupil and consider each ray as a source of the Huygens wavelet. Denoting the coordinates of ray on pupil plane is $(x', y', z')$ and on the sensor plane is $(x'', y'', z'')$, the complex amplitude on the sensor is superpositioned by the complex amplitude of spherical wavelet:
\begin{equation}
    \label{eq:coherent superposition}
    \mathbf{E}_{x''y''}(\mathbf{l}_{x'y'}, \mathbf{r}_{x''y''}, \mathbf{K}) = \sum_{y'}\sum_{x'}a_{0}\frac{e^{ik\mathbf{l}_{x'y'}}}{\mathbf{l}_{x'y'}}\frac{e^{ik\mathbf{r}_{x'y'}}}{\mathbf{r}_{x'y'}}\mathbf{K},
\end{equation}
where $l_{x'y'}$ is the optical path length from the source $(x_{0}, y_{0}, z_{0})$ to $(x', y', z')$. $k=2\pi / \lambda$ and $\lambda$ is the wavelength of the ray. $r_{x''y''}=(x''-x', y''-y', z''-z')$ indicates the direction of wavelet's propagation. $\mathbf{K}$ is the obliquity factor of wavelet, which is defined as follows:
\begin{equation}
    \label{eq:obliquity factor}
    \mathbf{K}(\mathbf{D}_{x'y'}, \mathbf{r}_{x''y''}, \mathbf{n}) = \frac{1}{2}[\cos{\langle \mathbf{n}, \mathbf{r}_{x''y''} \rangle}-\cos{\langle \mathbf{n}, \mathbf{D}_{x'y'} \rangle}],
\end{equation}
where $\mathbf{n}$ is the normal unit vector of the exit pupil plane and $\cos{\langle\cdot, \cdot\rangle}$ is the operation of computing the cosine value of the two vectors. The relationships of $\mathbf{n}$, $\mathbf{D_{x'y'}}$, and $\mathbf{r_{x''y''}}$ are magnified in \textbf{Fig.} \ref{fig:overview}. The complex amplitude is multiplied with its conjugate to obtain the intensity on sensor plane:
\begin{equation}
    \label{eq:intensity}
    I_{x''y''} = \mathbf{E}_{x''y''} \cdot \mathbf{E}^{*}_{x''y''},
\end{equation}

In this way, we obtain the PSF $I_{fovi}(\lambda)$ at different wavelengths of this FoV. The imaging simulation of edge is:
\begin{equation}
    \label{eq:imaging simulation}
    J_{e} = \int \mathcal{C}_{e}(\lambda)\cdot I_{fovi}(\lambda) d\lambda * L_{e} + N_{e}.
\end{equation}
here $\mathcal{C}_{e}(\lambda)$ is the sensor wavelength response. $J_{e}$, $L_{e}$, and $N_{e}$ are the observed edge, the latent ideal edge, and the sythetic noise image, respectively. We refer readers to \cite{10.1145/3474088} for details on imaging simulation implementation.

After simulating the degraded edge that resembles the actual observation, we measure the SFR as the procedure shown in \textbf{Fig.} \ref{fig:overview}b. First, we project all pixels along with the inclination to obtain the edge spread function (ESF). Second, a quarter of the pixel size $p_{i}$ is as the new sampling interval, and all pixel values that fall within the same sampling interval are averaged to represent the values of the resampling interval $i$. \textbf{This operation is the key to alleviate the influence of noise}. Third, we get LSF by the differentiation of ESF and calculate the discrete Fourier transformation of LSF. The normalized amplitude of the Fourier spectrum is the measured SFR. In this way, we construct a physical-based procedure to bridge the gap between the perturbed system parameters and the SFR measurement. 

\subsection{Proxy Camera Construction}\label{sec:3.2}
In this section, we present an optimization framework to construct a proxy camera so that its imaging simulation is similar to the photograph of the target device. As illustrated in Section. \ref{sec:3.1}, due to the highly nonlinear relationship between SFR and perturbed system parameters, it is impossible to predict the actual deviation of the camera analytically. So successive iterations are needed to approximate the solution. However, directly setting the SFR sequence as the target is unrealistic, where the computational overhead will increase exponentially when the sampling density grows. Therefore, SFR Area (SFRA), which is the area between the real measured SFR and the axis, is used as the target for optimization:
\begin{equation}
    \label{eq:real measured MTFA}
    SFRA^{*}_{fovi} = Area(SFR),
\end{equation}

We use the damped least-squares method to obtain the system solution of proxy camera \cite{4075580}. Let the perturbed parameters illustrated in Sec. \ref{sec:3.1.1} denoted by $p_{1}, p_{2}, \cdots, p_{N}$, where $N$ is the number of parameters. The simulated SFRA calculation can be represented by:
\begin{equation}
    \label{eq:MTFA calculation}
    SFRA_{fovi} = Area[F_{fovi}(p_{1}, p_{2}, \cdots, p_{N})],
\end{equation}
here $F_{fovi}$ indicates the operation to calculate SFR by the method detailed in Sec. \ref{sec:3.1}. The damped merit function $\phi(\Vec{P})$ in matrix-vector form is constructed as follows:
\begin{equation}
    \label{eq:merit function}
    \phi(\Vec{P}) = \Vec{f}^{T}\Vec{f} + \epsilon\Delta\Vec{P}^{T}\Delta\Vec{P},
\end{equation}
where $\epsilon$ is the damped factor. $\Delta\Vec{P}$ and $\Vec{f}$ are as follows:
\begin{equation}
    \label{eq:p and f in matrix-vector}
    \begin{split}
        \vec{P}_{k+1} &= (p^{k+1}_{1}, \cdots, p^{k+1}_{N})^\mathrm{T}, \vec{P}_{k} = (p^{k}_{1}, \cdots, p^{k}_{N})^\mathrm{T}, \\
        \Delta \vec{P}_{k} &= \vec{P}_{k+1} - \vec{P}_{k}, \qquad \quad \; \vec{f} = (f_{1}, f_{2}, \cdots, f_{M})^\mathrm{T},
    \end{split}
\end{equation}
$\vec{P}_{k+1}$ and $\vec{P}_{k}$ are the system parameters after $k$, $k+1$ iterations. $\Delta \vec{P}_{k}$ is the predicted linearity correction. $\vec{f}$ is the difference vector, where $f_{i} = SFRA^{*}_{fovi} - SFRA_{fovi}$ and $M$ is the number of sampled FoVs. According to the extreme value theory of multivariate function, the minimum of the merit function is achieved where its gradient is zero:
\begin{equation}
    grad\; \phi(\vec{P}) = A^\mathrm{T}\vec{f} + \epsilon\Delta\vec{P} = 0,
\label{eq:extreme value condition}
\end{equation}
here $A \in \mathbb{R}^{M\times N}$ is the partial derivative matrix ($A_{mn} = \partial f_{m} / \partial p_{n}$, where $m \in [1, M]$ and $n \in [1, N]$), which can be calculated by divided differences in implementation.

\begin{figure*}[htbp]
    \centering
    \includegraphics[width=\linewidth]{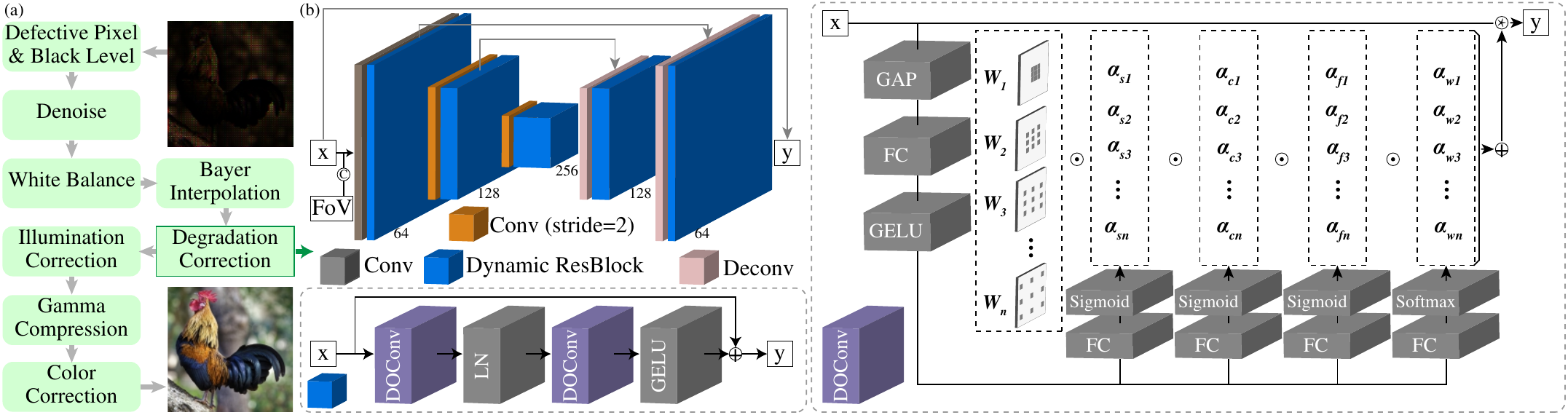}
    \caption{\textbf{Overview of the dynamic postprocessing pipeline}. (a) an optical degradation correction module is embedded into the ISP pipeline of mobile terminal. (b) we propose a dynamic postprocessing model based on dilated Omni-dimensional dynamic convolution, aiming at self-adaptively tackling the stochastic manufacturing deviation. All the layer configurations are marked with different colored blocks.}
    \label{fig:network}
\end{figure*}

Unfortunately, the $\vec{f}$ is a non-linear function where the variable $\vec{P}$ cannot be solved directly. Hence, linear approximation is made by the Taylor series of $\vec{f}$ at $\vec{P}_{k}$:
\begin{equation}
    \hat{f} = \vec{f}_{k} + A\Delta \vec{P}_{k},
\label{eq:talor approximation}
\end{equation}
where the $\vec{f}_{k}$ is the $\vec{f}$ of deviated parameters $\vec{P}_{k}$. We note that the approximation only guarantee linear accuracy in a small range of $\vec{P}_{k}$, so damping is set on $\Delta \vec{P}_{k}$ to control the step. Replacing the $\vec{f}$ in Eq. \eqref{eq:extreme value condition} by Eq. \eqref{eq:talor approximation}, we derive the $\Delta\vec{P}_{k}$ after $k$ iteration:
\begin{equation}
    \Delta\vec{P}_{k} = -(A^\mathrm{T}_{k}A_{k} + \epsilon I)^{-1}A^\mathrm{T}_{k}\vec{f}_{k}.
\label{eq:solution of delta p}
\end{equation}
here $A_{k}$ is the $A$ of $\vec{P}_{k}$, $I$ is the identity matrix. And the predicted system parameters are $\vec{P}_{k+1}=\vec{P}_{k}+\Delta\vec{P}_{k}$ after $k+1$ iteration. 

\begin{algorithm}[t]
    \caption{Proxy Camera Construction}
    \label{alg:alg1}
    \begin{algorithmic}[1]
        \REQUIRE {System parameters $\vec{P}_{0}$ , Measured $SFRA^{*}_{fovi}$}
        \STATE $k \gets 0$, and $\vec{f}_{k}, A_{k} \gets Merit(\vec{P}_{k}, SFRA^{*}_{fovi})$
        \WHILE{$||A^\mathrm{T}_{k}\vec{f}_{k}+\epsilon\Delta\vec{P}_{k}||\gg0$}
            \STATE $\Delta\vec{P}_{k}\gets-(A^\mathrm{T}_{k}A_{k}+\epsilon I)^{-1}A^\mathrm{T}_{k}\vec{f}_{k}$
            \STATE $\vec{P}_{k+1}\gets\vec{P}_{k}+\Delta\vec{P}_{k}$
            \STATE $k\gets k+1$, and $\vec{f}_{k}, A_{k} \gets Merit(\vec{P}_{k},SFRA^{*}_{fovi})$
        \ENDWHILE
        \RETURN $\vec{P}_{k}$
        \STATE \textbf{Function:} $Merit(\vec{P}_{k}, SFRA^{*}_{fovi})$
        \FOR{i=1:M}
            \STATE Edge imaging simulation with Eq. \eqref{eq:incident ray equation} - Eq. \eqref{eq:imaging simulation}
            \STATE Measure SFRA from synthetic edge as Fig. \ref{fig:overview}b
            \STATE $\vec{f}_{i}\gets SFRA^{*}_{fovi}-SFRA_{fovi}$
            \FOR{j=1:N}
                \STATE $A_{ij} = (\vec{f}(\cdots, p^{k}_{j}+\Delta p_{j}, \cdots) - \vec{f}_{k}) / \Delta p_{j}$
            \ENDFOR
        \ENDFOR
        \RETURN $\vec{f}_{k}$, $A_{k}$
    \end{algorithmic}
\end{algorithm}

We also note that each system parameter contributes differently to the merit functions. Therefore, we dynamically adjust the damping factor $\epsilon$ of each parameter according to the non-linearity of the solution $\Delta \vec{P}$, which is detailed in the \textbf{supplementary file}. In this way, the optimization is performed from \textbf{Fig.} \ref{fig:overview}a to \textbf{Fig.} \ref{fig:overview}c until the gradient of solution (Eq. \eqref{eq:extreme value condition}) is close to zero.

Altogether, we illustrate the perturbed optical system model and the method to construct a proxy camera based on the measurement of SFR. \textbf{There are three significant advantages between the proposed method and other end-to-end optimizing approaches in Section \ref{sec:related area1}. First, our optimization framework is insensitive to noise because of the resampling in SFR measurement. Second, our imaging simulation considers the diffraction effect caused by the optical aperture and therefore is more accurate than bare ray tracing. Third, we consider the potential perturbation of all elements which is more authentic in following the physical procedure} (in-depth comparisons are presented in \textbf{supplementary file}). Hence, we engage the proxy camera with the physical formation pipeline of raw images to perform authentic imaging simulation, generating the data pairs for deep-learning-based reconstruction.

\section{Dynamic-processing ISP}\label{sec:4}

To eliminate the camera-wise manufacturing deviations and spatially varying optical aberrations, the ability of self-adaptive correction is necessary for the postprocessing pipeline. Moreover, the computational overhead of the mobile terminal puts a significant limit on the complexity of the model. To this end, we propose a lightweight framework based on dynamic convolution to meet the needs of adaptive processing and the constraints in application. In the following Section. \ref{sec:4.1}, we first illustrate the data preparation for the training of the framework. Then in the Section. \ref{sec:4.2}, the proposed postprocessing pipeline is detailed.

\subsection{Data Preparation}\label{sec:4.1}
Based on the virtual camera constructed in Sec. \ref{sec:3}, we construct the training data pairs by the imaging simulation in Sec. \ref{sec:3.1.2}. Since the exposure parameters of real photography are discrete, the dynamic range of the captured raw image is not ideal. Therefore, we add luminosity compression/decompression in simulation which is different from the transformation in \cite{10.1145/3474088}. After obtaining the raw-like image, the formation of the sensor observation with optical degradation can be formulated as:
\begin{equation}
    \label{eq:optical degraded observation}
    J_{e}(x, y) = \int \mathcal{C}_{e}(\lambda)\cdot I_{fovi}(x, y, \lambda) d\lambda * L_{e}(x, y) + N_{e},
\end{equation}
here $(x, y)$ is the pixel coordinates on the sensor plane. And the rest denotions are the same as Eq. \eqref{eq:imaging simulation}. We note that the performance of optical degradation on the raw image is linear and channel-irrelevant. So different from \cite{10.1145/3474088}, we abandon the CCM and the gamma compression in the synthetic pipeline after adding the degradation into the image. Because these operations will introduce non-linearity and cross-channel information, thus increasing the difficulty of restoration. Because of the limited number of manufacturing samples, the proxy cameras could not cover the distribution of manufacturing deviation. So we regard the max bias of each system parameter as the tolerance and random sample them to generate more virtual optical systems for data generation. This augmentation allows the model to learn the potential degradation and prevents overfitting. In this way, we obtain the RAW-to-sRGB data pairs, which characterize the mapping of optical degradation and encode the accumulated errors of cascaded modules. The data pairs not only follow the physical procedure of image formation but also are friendly to the training of the deep-learning method.

\subsection{Dynamic Postprocessing based on DOConv}\label{sec:4.2}

The stochastic manufacturing deviations and spatially varying optical aberrations cause different PSFs on the sensor plane. However, the traditional convolution operation strictly receives the feature by fixed weight and locations around its center, which is hard to adapt to the stochastic degradation and introduce relevant information into the output. To this end, we propose a dynamic convolution model for eliminating the optical degradation and embed it in the traditional ISP pipeline to realize extreme quality computational imaging. Inspired by the idea in \cite{li2021omni}, we design the dilated Omini-dimensional dynamic convolution (DOConv) and implement it into a variant of UNet architecture. As shown in \textbf{Fig.} \ref{fig:network}, each DOConv has four weights, and their dilations vary from 1 to 4 in implementation, aiming at performing targeted feature acquisition. We reduce the stages of the architecture to mitigate the computational overhead. And the ResBlocks, whose internal components are the same as the Block in \cite{liu_convnet_2022}, are applied in each scale to enhance the expression ability. The model takes degraded raw-like data as input and outputs restored images in the same domain. Subsequently, the restorations are processed by the subsequent modules and supervised by the sRGB ground-truth. Since the generated data pairs are pixel-to-pixel aligned and cover all potential degrading distributions, it is sufficient to train the model only relying on fidelity losses:
\begin{equation}
    \label{eq:loss function}
    L(\theta) = \frac{1}{N}\sum^{N}_{n=1}||Process(Model(J^{n}_{e})) - L^{n}||^{2}_{2}.
\end{equation}
where $\theta$ denotes the learned parameters in the model. $J^{n}_{e}$ are the degraded raw image and $L^{n}$ are the corresponding sRGB ground-truths. $Process(\cdot)$ denotes the subsequent operations after optical degradation correction.

\section{Experiments}\label{sec:5}
\begin{figure*}[htbp]
    \centering
    \includegraphics[width=\linewidth]{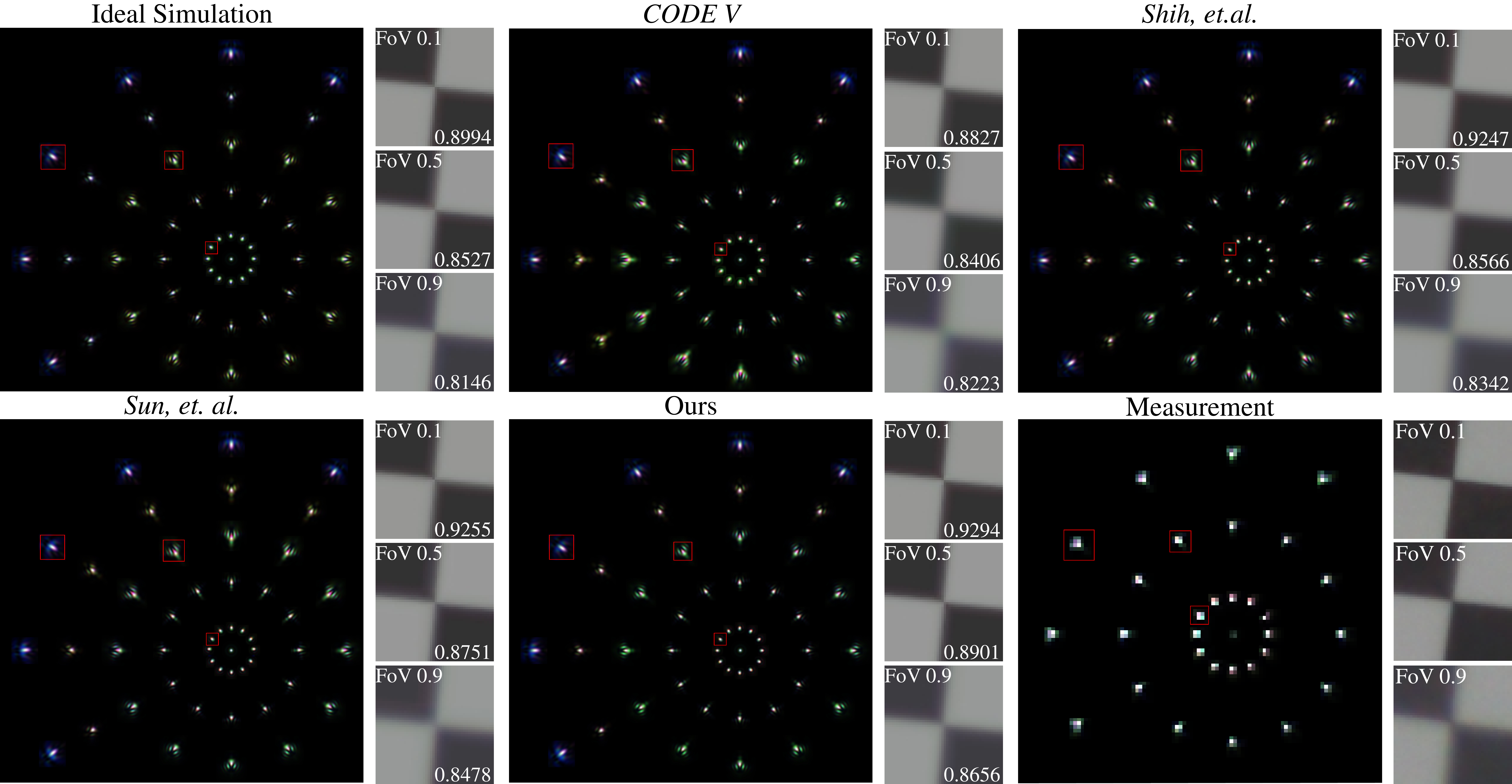}
    \caption{\textbf{Qualitative evaluation on PSF and imaging simulation}. We visualize the centrosymmetric PSFs ($10\times$resampling for detailed comparisons) of the proxy cameras constructed from one machining sample of the Phone (the measured PSFs are shown in the sensor resolution). For the proxy camera, the resolutions of PSFs are 50, 50, 80, 80, and 120 of FoV 0.1, 0.3, 0.5, 0.7, and 0.9, respectively. And in measurement, the resolutions of PSFs are 5, 8, and 12 for FoV 0.1, 0.5, and 0.9. We present the imaging simulation results and their SSIM compared with actual measurements.}
    \label{fig:deviated psf}
\end{figure*}

We first roughly illustrate the experimental setting in Section. \ref{sec:5.1}. In Section \ref{sec:5.2}, comprehensive experiments are conducted to demonstrate the theoretical advantage of the proposed proxy camera construction. In Section \ref{sec:5.3}, we evaluate the strength of the proposed dynamic model when tackling optical degradation. Finally, an in-depth ablation study is presented in Section \ref{sec:5.4}.

\subsection{Experimental setting}\label{sec:5.1}
To substantiate the authenticity of the proposed proxy camera construction, we evaluate two devices, one is a customized DSLR camera, and another one is Huawei Honor 20 Pro (Phone). The optical prescriptions of both cameras are known, and their system parameters are listed in the \textbf{supplementary file}. In edge measurement (\textbf{Fig.} \ref{fig:overview}a), we rotate the targets at $9 \sim 12^{\circ}$ angles and take photos of them. And the ideal edge is colored according to the dynamic range of the measured edge. For SFRA measurement (\textbf{Fig.} \ref{fig:overview}b), the sampled FoVs are the regions that evenly divide the image into $15\times 20$. In training data construction, we first calculate the PSFs of the proxy camera following Eq. \eqref{eq:intensity}. Then the PSFs of different FoVs are used to degrade the latent image as Eq. \eqref{eq:imaging simulation}. We acquire the latent image by adopting DIV2K \cite{martin2001database} and rescaling these data to the resolution of the camera (DSLR is $4000\times 6000$ and Phone is $3000\times 4000$). In terms of the hyperparameters of training, the channel of each layer is marked at the block bottom, the model is trained with ADAM optimizer \cite{kingma2014adam} ($\beta_{1}=0.9$, $\beta_{2}=0.999$, $\epsilon=10^{-8}$), and the learning rate starts at $10^{-4}$ then halved every 10 epochs. The setting of dynamic convolution is the same as \cite{li2021omni}. For more implementation details about the proposed framework, please refer to the \textbf{supplementary file}.

\subsection{Authenticity of Proxy Camera Construction}\label{sec:5.2}
\subsubsection{Competing Methods}\label{sec:5.2.1}
To demonstrate the advantages of the proposed approach, we compare our method with these representative methods
\begin{enumerate}
    \item The built-in tolerance analysis of optical design software, \textit{i.e.}, MTF tolerance of $CODEV^{\circledR}$ \cite{rimmer_tolerancing_1978}.
    \item Modify the optical parameters by the calibrated PSFs, \textit{i.e.}, \textit{shih, et. al.} \cite{hutchison_image_2012}.
    \item Optimize the system by image-to-image rendering through differentiable ray tracing, \textit{i.e.}, \textit{sun, et. al.} \cite{sun_end--end_2021}.
\end{enumerate}
Since the tolerancing in commercial software only derives the deviation range, we choose the prediction's median as the bias for each parameter. In the second method, we use pixel-level MSE between the simulated and measured PSFs to guide the optimization. Noting that there are alternative end-to-end system optimization besides the third method, we present an in-depth comparison in the \textbf{supplementary file} to illustrate the reason we pick this method.


\subsubsection{Quantitative and Qualitative Assessment}\label{sec:5.2.2}

\begin{table}[t]
    \caption{\textbf{Quantitative proximity between real devices and proxy camera}. Note that the value is the average of all machining samples. The best and the second-best indicators of each FoV are marked in \textcolor{red}{red} and \textcolor{blue}{blue}.}
    \label{tab:quantitative comparisons of proxy}
    \centering
    \begin{tabular}{l c c c}
        \hline
        \multirow{2}{*}{CAMERA} & \multirow{2}{*}{Method} & MSE ($\times10^{-3}$) $\downarrow$      & SSIM $\uparrow$           \\
        \cline{3-4}
                                &                         & FoV 0.1/0.5/0.9            & FoV 0.1/0.5/0.9 \\
        \hline
        \hline
        \multirow{4}{*}{DSLR}   & Ideal                   & 0.576/0.804/1.023          & 0.913/0.906/0.872 \\
        \cline{2-4}
                                & $CODEV^{\circledR}$     & 0.521/0.764/1.242          & 0.899/0.874/0.833 \\
        \cline{2-4}
                                & \textit{shih, et. al.}  & \textcolor{red}{0.374}/0.714/0.958          & \textcolor{red}{0.951}/\textcolor{blue}{0.932}/0.917 \\
        \cline{2-4}
                                & \textit{sun, et. al.}   & 0.407/\textcolor{blue}{0.708}/\textcolor{blue}{0.924}          & 0.942/0.929/\textcolor{blue}{0.920} \\
        \cline{2-4}
                                & Ours                    & \textcolor{blue}{0.382}/\textcolor{red}{0.698}/\textcolor{red}{0.877}          & \textcolor{blue}{0.946}/\textcolor{red}{0.938}/\textcolor{red}{0.922} \\
        \hline
        \hline
        \multirow{4}{*}{Phone}  & Ideal                   & 0.814/1.621/1.768          & 0.896/0.854/0.806 \\
        \cline{2-4}
                                & $CODEV^{\circledR}$     & 0.852/1.746/1.496          & 0.878/0.842/0.822 \\
        \cline{2-4}
                                & \textit{shih, et. al.}  & 0.643/1.386/1.325          & 0.913/0.858/0.832 \\
        \cline{2-4}
                                & \textit{sun, et. al.}   & \textcolor{blue}{0.628}/\textcolor{blue}{1.303}/\textcolor{blue}{1.157}          & \textcolor{blue}{0.915}/\textcolor{blue}{0.872}/\textcolor{blue}{0.847} \\
        \cline{2-4}
                                & Ours                    & \textcolor{red}{0.547}/\textcolor{red}{1.251}/\textcolor{red}{0.952}          & \textcolor{red}{0.927}/\textcolor{red}{0.887}/\textcolor{red}{0.859} \\
        \hline
    \end{tabular}
\end{table}
To illustrate the authenticity of the proposed optimization, we evaluate the proximity of the proxy camera (by the MSE between the simulated and the measured discrete SFR) and the similarity of imaging (by SSIM \cite{1284395} between images). The indexes of all machining samples are averaged for comparison, and the ideal simulations are provided for reference. As shown in \textbf{Tab.} \ref{tab:quantitative comparisons of proxy}, the competing methods generally perform better on the DSLR than on the Phone. These phenomena attribute to 1) the aberration of DSLR increases uniformly with the growth of FoV, yet the degradation of Phone changes significantly, which increases the difficulty in prediction; 2) the relative illumination of PHONE decreases a lot in marginal FoV, resulting in a declining Signal-to-Noise ratio (SNR). So for the optimizing method that relies on pixel-level indicators (\textit{shih, et. al.} and \textit{sun, et. al.}), it is challenging to extract undisturbed information from actual noisy measurements. In terms of evaluation metrics, our method constructs the proxy systems that are the closest to the machined devices, especially in difficult situations such as low SNR and highly non-uniformed degradation.

The visualization of PSF calculation and the imaging simulation of all competing methods are shown in \textbf{Fig.} \ref{fig:deviated psf}. Limited by the space, we only present the resampling PSFs and imaging simulation of some FoVs. Moreover, the ideal simulation and the actual measurement are provided for reference, and the SSIM of each simulated patch is listed for additional quantitive evaluation. As shown by the PSFs of the same FoV, all methods roughly predict the eccentricity of the device. We note that the degradation caused by the predicted deviations of $CODEV^{\circledR}$ is more severe than the actual measurement, which is because it calculates the influence of each system variable separately without considering the combined impact of the entire lens. \textit{shih, et. al.} and \textit{sun, et. al.} have poor accuracy when the FoV increases due to the pixel-level metric disturbance introduced by the magnified noise level in edge FoV. \textit{sun, et. al.} obtains more accurate simulation results than $CODEV^{\circledR}$ and \textit{shih, et. al.}. But since the ray tracing does not consider the diffraction of optical aperture, its predicted perturbation is often larger than our estimation, resulting in more significant degradation. By contrast, our method produces accurate results not only on visual similarity but also from quantitative assessment. More discussions are presented in the \textbf{supplementary file}.

\subsubsection{Noise Injection Influence}\label{sec:5.2.4}
\begin{figure*}[ht]
    \centering
    \includegraphics[width=\linewidth]{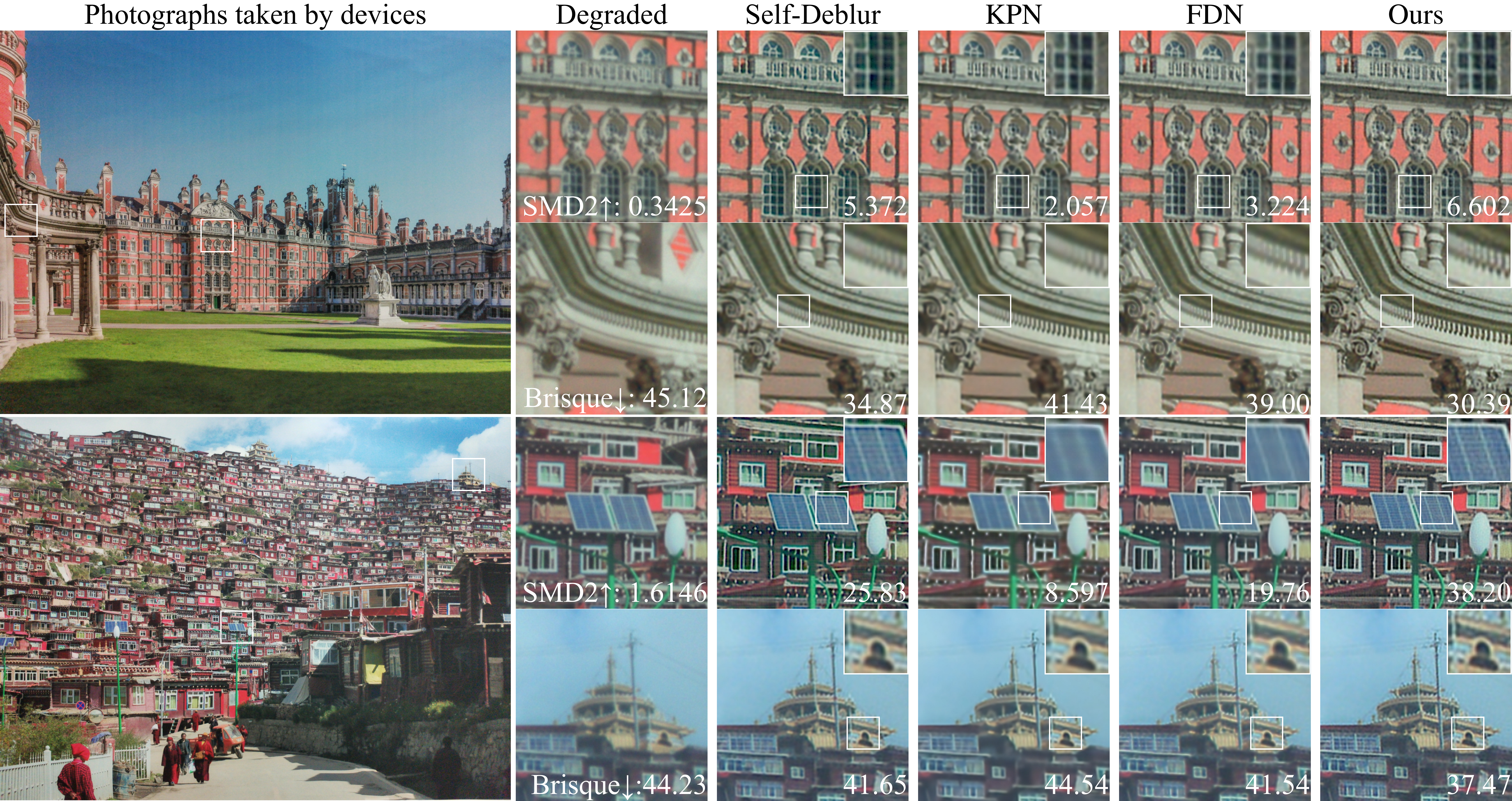}
    \caption{\textbf{Non-uniform deblurring on real photographs of Phone}. We mark the corresponding regions with white boxes and present the indicators.}
    \label{fig:non uniform comparisons}
\end{figure*}

\begin{figure}[t]
    \centering
    \includegraphics[width=\linewidth]{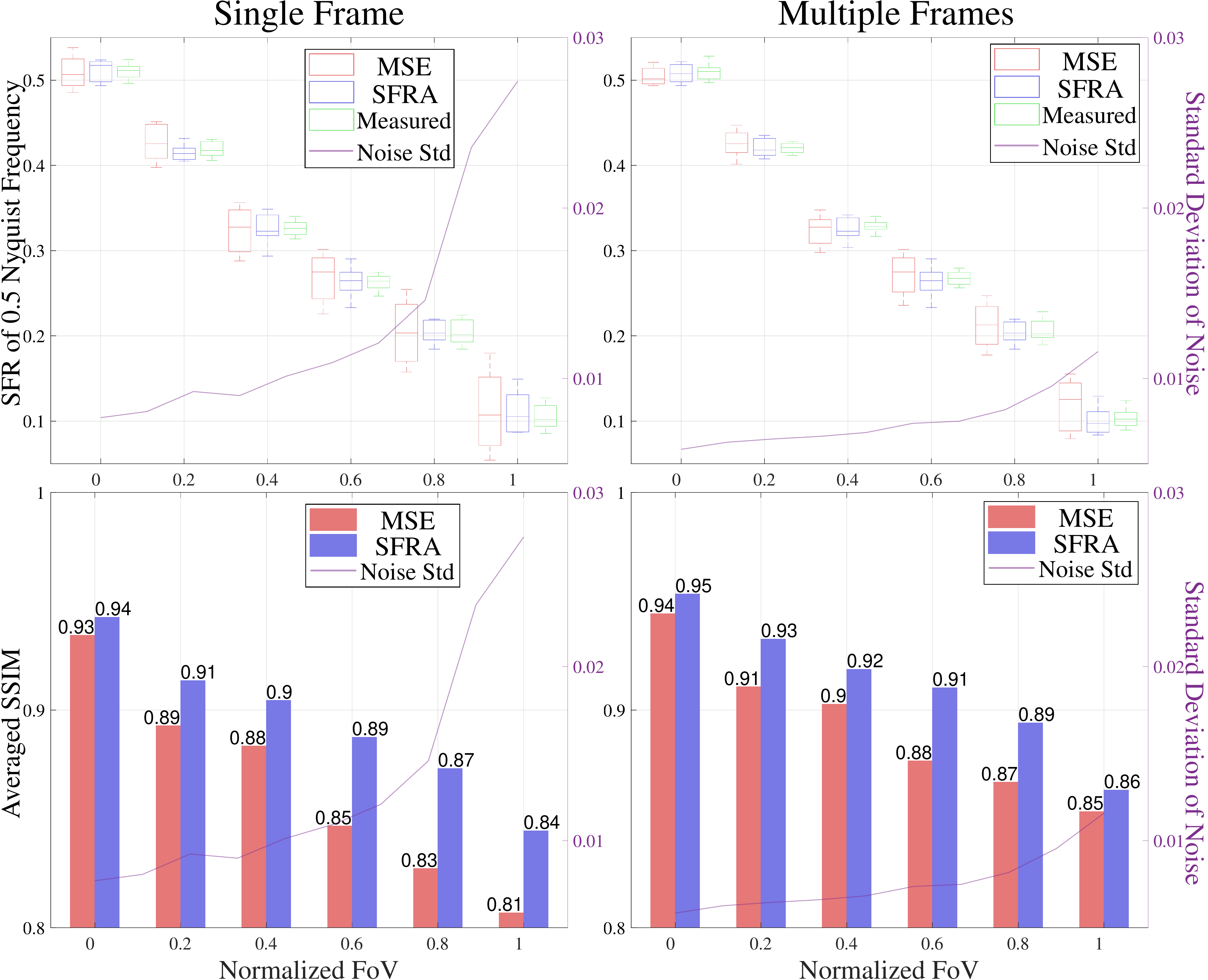}
    \caption{\textbf{Performance under different noise level}. The left/right part shows the accuracy of proxy camera construction under single/multiple frames. The upper/lower part presents the MTF/SSIM evaluation. The noise level is plotted in the purple line.}
    \label{fig:noise injection}
\end{figure}

The crucial merit of our approach compared to optimizing with pixel-level evaluation is that the measuring procedure of SFRA is insensitive to the noise of actual photographs. We analyze the statistical results when using pixel-level MSE and SFRA as optimization objectives under different noise situations. The left part and the right part of \textbf{Fig.} \ref{fig:noise injection} show the different FoVs' accuracy on the single-frame image and the multi-frames image (superposed for denoising), respectively. The evaluation in the upper part of \textbf{Fig.} \ref{fig:noise injection} is the SFR at 0.5 Nyquist frequency, and the lower part is the SSIM between the simulation and the actual photograph. We note that the variance of noise (indicated by the purple line) grows as the FoV increases because of the decreasing illumination by lens shading. Thus the optimization targeted with MSE is prone to severe fluctuations due to this metric being easily affected by accidental errors when the noise level increases. On the contrary, the proposed method maintains similar-to-real SFR fluctuation and higher SSIM evaluation under different noise levels. This mainly benefits from the resampling ESF operation (projecting all pixels along with the inclination and averaging the values), which makes the SFRA insensitive to the noise and provides stable guidance for our optimization.

\subsection{Evaluation on Dynamic Postprocessing}\label{sec:5.3}
\subsubsection{Competing Methods}\label{sec:5.3.1}
The performance of optical degradation on the image can be integrated as the influences of spatially varying blur on different FoVs. Therefore, correcting the optical degradation can be summarized as a deblurring task. we collect the state-of-the-art deblurring algorithms for comparisons
\begin{enumerate}
    \item Global deblurring method, \textit{i.e.}, scale recurrent network (SRN) \cite{tao2018scale} and Self-Deblur (SD) \cite{ren2020neural}.
    \item Kernel-based deblurring method, \textit{i.e.}, kernel prediction network (KPN) \cite{guo2020efficientderain}.
    \item Dynamical-adjusted deblurring method, \textit{i.e.}, FoV deformable network (FDN) \cite{10.1145/3474088}.
\end{enumerate}
For these algorithms, we apply the same RAW-to-sRGB data for evaluation. The details of training and inference procedure are presented in the \textbf{supplementary file}.

\subsubsection{Results on non-uniform deblurring}\label{sec:5.3.2}

\begin{table*}[ht]
    \caption{\textbf{Quantitative results on sythetic data and real photographs}. \textbf{C} denotes the evaluated platform. \textbf{T} is the training datasets, where single sample means the data pairs are synthetic by one proxy camera and multiple samples means altogether evaluation on multiple virtual cameras. The best and the second-best indicators of each evaluation are marked in \textcolor{red}{red} and \textcolor{blue}{blue}. We indicate the better metrics with up/down arrows.}
    \label{tab:quantitative of reconstruction}
    \setlength{\tabcolsep}{1.4mm}
    \centering
    \begin{tabular}{c c c c c c c c c c c c c c c c}
        \hline
        \multirow{2}{*}{C} & \multirow{2}{*}{T} & \multirow{2}{*}{Method} & \multicolumn{4}{c}{Evaluation on Synthetic Data} & \multicolumn{2}{c}{Real Photographs} & \multirow{2}{*}{T} &  \multicolumn{4}{c}{Evaluation on Synthetic Data} & \multicolumn{2}{c}{Real Photographs} \\
        \cline{4-9} \cline{11-16}
                                 &                         &                         & PSNR $\uparrow$  & SSIM $\uparrow$  & VIF $\uparrow$   & LPIPS $\downarrow$ & BRISQUE $\downarrow$ & NIQE $\downarrow$ &   & PSNR $\uparrow$  & SSIM $\uparrow$  & VIF $\uparrow$   & LPIPS $\downarrow$ & BRISQUE $\downarrow$ & NIQE $\downarrow$ \\
        \hline
        \hline
        \multirow{5}{*}{\rotatebox{90}{DSLR}} & \multirow{10}{*}{\rotatebox{90}{Single Sample}} & SNR                     & 42.19 & 0.984 & 0.988 & 1.383 & 39.80 & 3.846 & \multirow{10}{*}{\rotatebox{90}{Multiple Samples}}    & 40.84 & 0.973 & 0.967 & 1.839 & 42.82 & 4.153 \\
                                 &                         & SD                      & 42.58 & 0.988 & 0.989 & 1.178 & \textcolor{blue}{38.51} & 3.258 &    & 40.97 & 0.976 & 0.970 & 1.678 & \textcolor{blue}{40.88} & 3.868 \\
                                 &                         & KPN                     & 41.86 & 0.987 & 0.982 & 1.186 & 43.28 & 3.729 &    & 40.57 & 0.977 & 0.966 & 1.523 & 45.86 & 4.257 \\
                                 &                         & FDN                     & \textcolor{red}{43.12} & \textcolor{red}{0.992} & \textcolor{red}{0.996} & \textcolor{red}{0.757} & 40.57 & \textcolor{red}{2.924} &    & \textcolor{blue}{41.94} & \textcolor{blue}{0.985} & \textcolor{blue}{0.987} & \textcolor{red}{1.212} & 42.89 & \textcolor{blue}{3.505} \\
                                 &                         & Ours                    & \textcolor{blue}{43.03} & \textcolor{blue}{0.991} & \textcolor{blue}{0.993} & \textcolor{blue}{0.924} & \textcolor{red}{36.56} & \textcolor{blue}{3.213} &    & \textcolor{red}{42.12} & \textcolor{red}{0.986} & \textcolor{red}{0.990} & \textcolor{blue}{1.241} & \textcolor{red}{38.88} & \textcolor{red}{3.431} \\
        \cline{1-1} \cline{3-9} \cline{11-16}
        \multirow{5}{*}{\rotatebox{90}{Phone}} &          & SNR                      & 33.26 & 0.957 & 0.943 & 2.264 & 43.27 & 4.674 &    & 30.17 & 0.928 & 0.907 & 3.248 & 47.91 & 5.347 \\
                                 &                         & SD                      & 34.03 & 0.959 & 0.946 & 1.966 & \textcolor{blue}{41.77} & 4.185 &    & 30.33 & 0.930 & 0.913 & 3.017 & \textcolor{blue}{45.86} & 4.928 \\
                                 &                         & KPN                     & 32.96 & 0.963 & 0.939 & 1.525 & 44.66 & 4.527 &    & 29.96 & 0.938 & 0.901 & 2.476 & 48.57 & 5.267 \\
                                 &                         & FDN                     & \textcolor{red}{34.56} & \textcolor{blue}{0.974} & \textcolor{red}{0.968} & \textcolor{red}{0.984} & 42.71 & \textcolor{red}{3.644} &    & \textcolor{blue}{31.86} & \textcolor{blue}{0.942} & \textcolor{blue}{0.922} & \textcolor{blue}{1.897} & 46.59 & \textcolor{red}{4.056} \\
                                 &                         & Ours                    & \textcolor{blue}{34.28} & \textcolor{red}{0.976} & \textcolor{blue}{0.964} & \textcolor{blue}{1.084} & \textcolor{red}{39.88} & \textcolor{blue}{3.713} &    & \textcolor{red}{32.28} & \textcolor{red}{0.950} & \textcolor{red}{0.938} & \textcolor{red}{1.584} & \textcolor{red}{43.68} & \textcolor{blue}{4.131} \\
        \hline
    \end{tabular}
\end{table*}
\begin{figure*}[t]
    \centering
    \includegraphics[width=\linewidth]{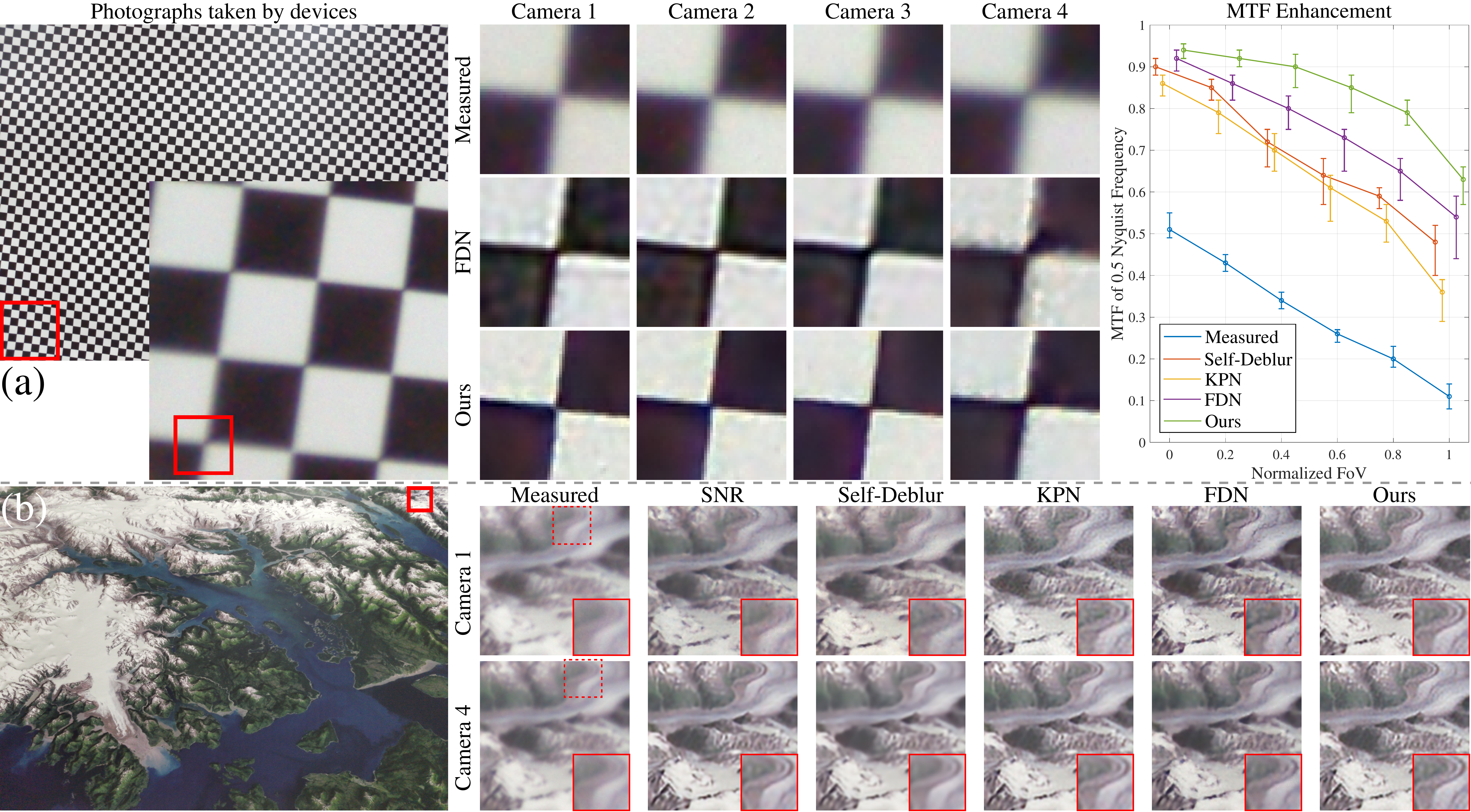}
    \caption{\textbf{Adaptation for manufacturing deviation}. (a) we show the magnified patch to illustrate the image quality mutation of Phone. The actual checkers' restoration of different machining samples are present for comparison. And we evaluate the average SFR (MTF) enhancements on the machining samples of test set. (b) the natural photograph restoration when applied on different machining samples of Phone.}
    \label{fig:camera wise comparing}
\end{figure*}

To evaluate the ability of various methods to handle non-uniform optical degradation, we train all the compared methods on the data generated by one proxy camera. \textbf{Fig.} \ref{fig:non uniform comparisons} shows the various methods' restoration in different FoVs of real photographs and presents the BRISQUE \cite{6272356} and SMD2 \cite{krotkov_focusing_1988} for evaluation. The quantitative evaluation of non-uniform deblurring is provided in \textbf{Tab.} \ref{tab:quantitative of reconstruction}, where the single sample of \textbf{T} is the assessment on one proxy camera. We apply the reference image quality assessments (IQA), \textit{e.g.}, PSNR, SSIM \cite{1284395}, VIF \cite{1576816}, and LPIPS \cite{zhang2018unreasonable} to evaluate synthetic data. And the non-reference IQAs, \textit{e.g.}, BRISQUE \cite{6272356}, NIQE \cite{mittal2012making} are assessing real photograph because obtaining non-degraded reference of real photo is impossible.

For the globally consistent deblurring method, it can be seen that the deterioration is suppressed. However, since the optical aberration is spatially variant, the restoration is a compromise of each FoV: in the center FoV with less blurring, ringing, and color artifacts shows up on the border of objects, while in the case of severe degradation at the edge of the image, the optical aberration is not fully corrected. When it comes to the kernel-based method, KPN can cope with the non-uniform degradation, benefiting from the ability to predict the spatially varying kernel according to the input features. However, since the blurring is associated with the FoV and these models are not aided by FoV information, such methods are confused by the input features and generate incorrect restoration results. For the approach that adjusts processing according to the FoV, FDN can better tackle the FoV-related optical aberration and obtain competitive results when correcting a specific camera (shown in \textbf{Tab.} \ref{tab:quantitative of reconstruction}). Our model realizes competitive results in both visual quality and metrics. We note that we outperform the FDN in BRISQUE, which may be because this metric pays more attention to the details of the image. The dynamic convolution model generally suffers from overfitting when the data distribution is relatively singular. But our method fits the degradation of the actual measurement well when only depending on the synthetic data of one proxy camera, which may be due to the dilation manner so that different weights have different specializations for adaptive processing. 

\subsubsection{Adaptation for Manufacturing Deviation}\label{sec:5.3.3}

In mass production, there is no time to specifically train each post-processing model for a particular camera. To this end, the restoration must be able to cope with the stochastic deviation introduced in manufacturing and assembly. To evaluate the ability to tackle this task, we simultaneously train the competing methods with the data generated by various virtual cameras. \textbf{Fig.} \ref{fig:camera wise comparing}a and \textbf{Fig.} \ref{fig:camera wise comparing}b show the results of different models dealing with the camera-wised degradation on the same scene. The measurement deteriorates more unpredictably due to random manufacturing deviation when the FoV increases, which enlarges the difficulty of restoration. Comparing the experimental results, we note that the static model (weights are fixed after completing training) cannot fit well in the diverse deviation of real manufacturing. In consequence, the static model fails to tackle the different machining degradations between real cameras, especially when FoV increases. Benefiting from the dynamic convolution, our model adaptively restores the degraded features and achieves better restoration results on each camera. The SFR enhancement (shown in the right part of \textbf{Fig.} \ref{fig:camera wise comparing}a) demonstrates that the proposed method realizes better and more stable restoration. Other methods suffer from significant fluctuation when FoV increases. Additional quantitative evaluation of deviation adaptation presents in \textbf{Tab.} \ref{tab:quantitative of reconstruction}. Other competing methods receive unfavorable results when the data distribution becomes complicated, while the proposed method maintains a high level.


\subsection{Ablation Study}\label{sec:5.4}
\begin{figure}[t]
    \centering
    \includegraphics[width=\linewidth]{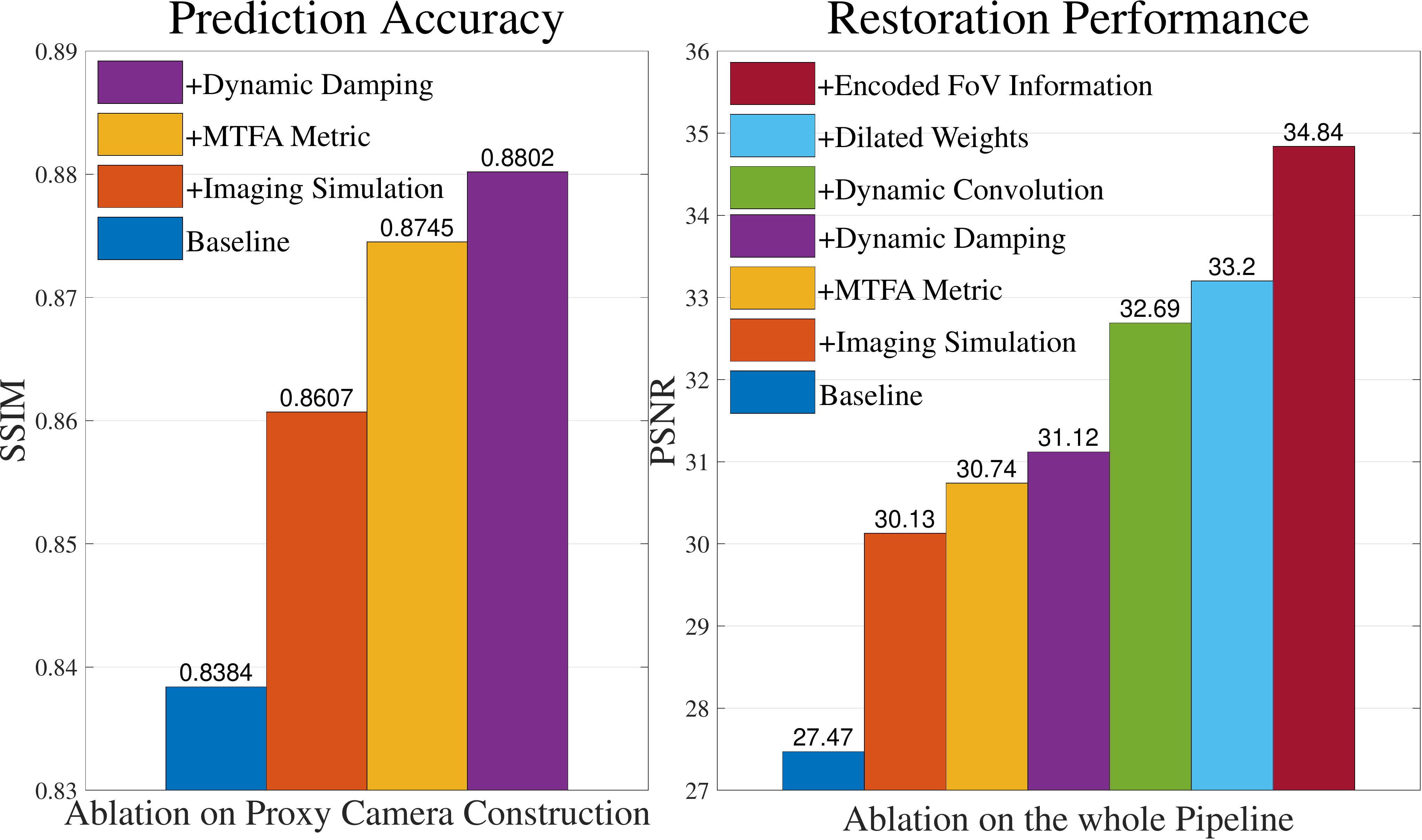}
    \caption{\textbf{Ablation study on the proxy camera construction} (accuracy evaluation on the SSIM between the simulated images and the natural photographs). And \textbf{Ablation study on the whole pipeline} (restoration assesment on the test synthetic data sets).}
    \label{fig:ablation study}
\end{figure}
We first evaluate the prediction accuracy when ablating the crucial modules of virtual camera prediction. Then comprehensive ablation studies are conducted on our proposed method. Specifically, for the proxy camera construction: 1) we ablate the imaging simulation and calculate MTF by the continuous wavefront at the pupil plane. 2) we replace the MTFA with the MTF at 0.5 Nyquist frequency. 3) we substitute the dynamic damping strategy with a fixed damping factor. For the dynamic restoration: 1) the dynamic convolution layers are replaced by the ordinary convolution. 2) the dilation of each weight is the same; 3) encoded FoV information is ablated and only inputting image feature. 

As shown in \textbf{Fig.} \ref{fig:ablation study}, the accuracy of prediction is significantly affected after ablating the imaging simulation module, which is mainly due to the inherent difference between the MTF calculation. One way is to compute by the continuous wavefront at the pupil plane, and another way is to obtain the SFR of the edge image (as demonstrated in the \textbf{supplementary file}). Moreover, using SFRA for optimization alleviates the influence of the measured SFR's singular value in some spatial frequencies. And the dynamic damping ensures that the solution of perturbed parameters is linear within the damping range. Both of these modules facilitate the proposed framework more stable and efficient. For the evaluation of image restoration, the improvements in prediction accuracy are positively relevant. In the ablation of the reconstruction model, dynamic convolution performs a better fitting than traditional convolution when the distribution of data is more diverse. Meanwhile, the different dilations of weights also facilitate the correction of spatially-variant degradation. Finally, the encoded FoV information has a positive gain on restoration since the degradation is strongly associated with the pixel position on the sensor.

\begin{figure}[t]
    \centering
    \includegraphics[width=\linewidth]{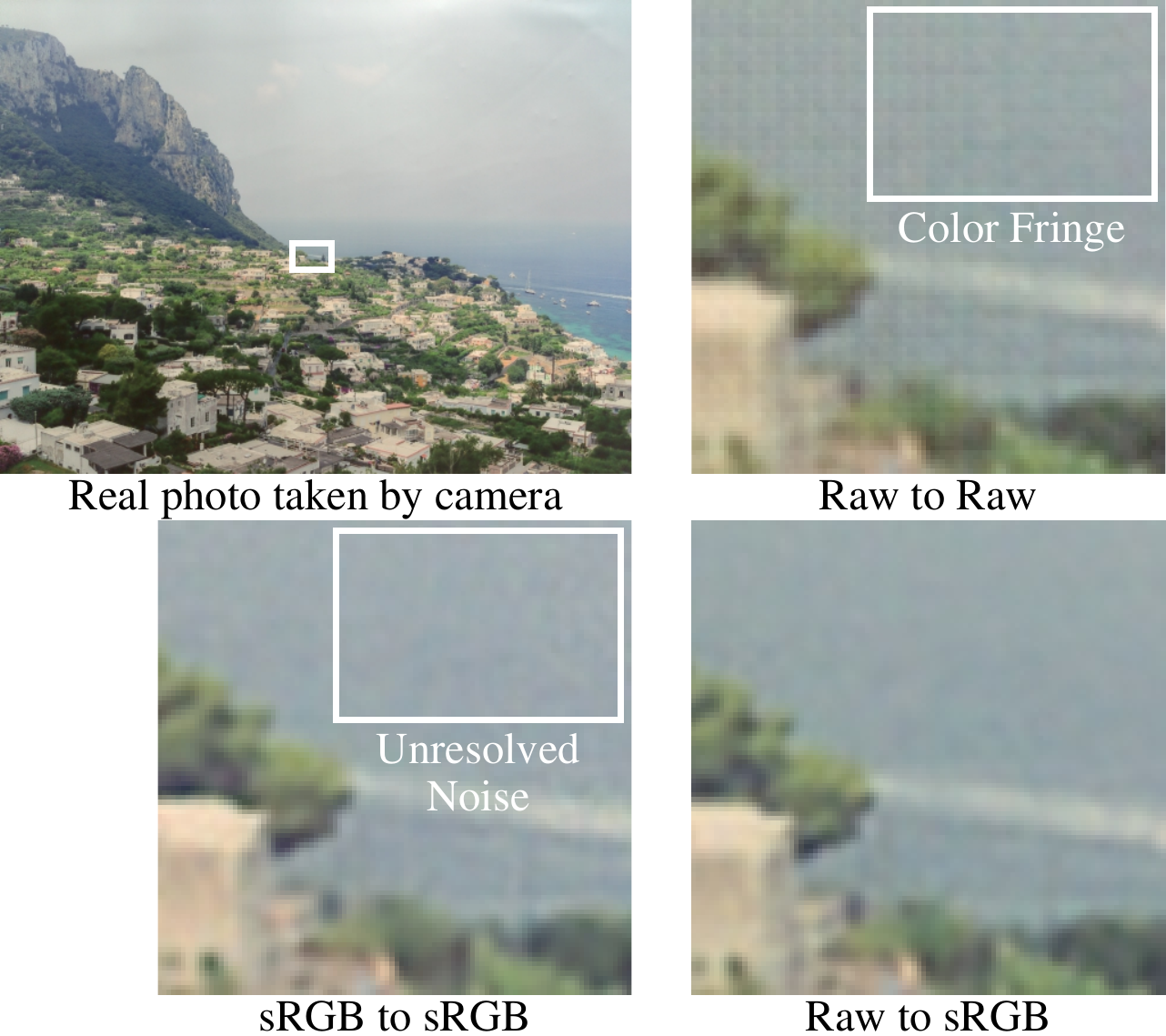}
    \caption{\textbf{Ablation on the dynamic processing ISP}. We evaluate the performance when placing the optical degradation correction model at different positions of ISP pipeline.}
    \label{fig:raw2rgb}
\end{figure}

\begin{figure*}[ht]
    \centering
    \includegraphics[width=\linewidth]{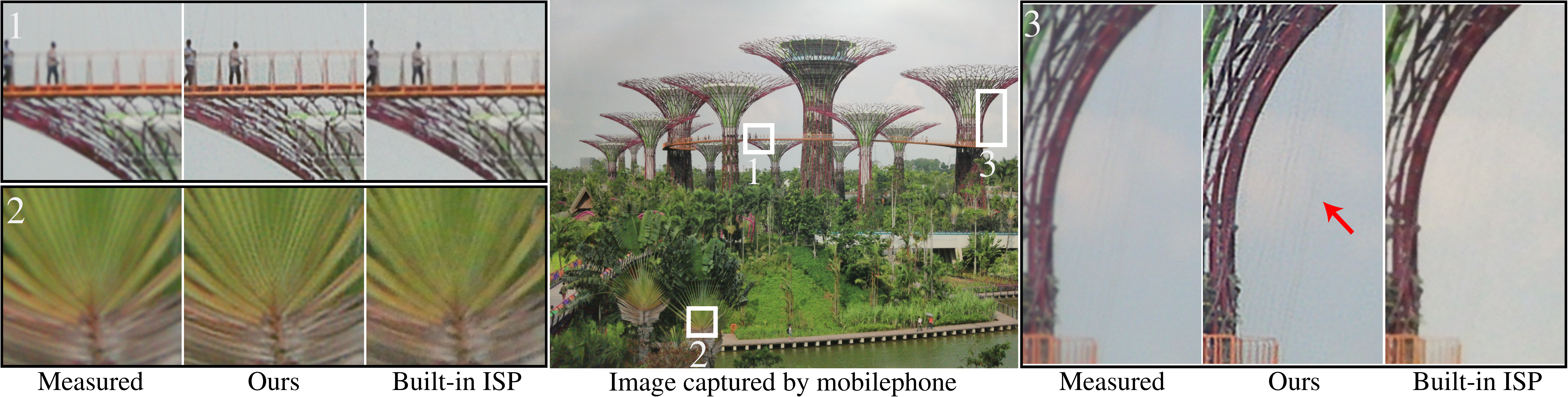}
    \caption{\textbf{Comparisons with built-in ISP}. Note that the built-in ISP smooths out the cords, yet our method restores it (marked with the red arrow).}
    \label{fig:isp comparisons}
\end{figure*}

In the proposed dynamic-processing ISP, we deploy the aberration recovery module between the Bayer interpolation and luminance correction. Its output is manipulated by the subsequent modules and compared with sRGB ground-truth in training. To prove the rationality of this deployment, we conducted the following ablation experiments: 1) Construct raw-to-raw data without considering subsequent operations for training; 2) Deploy the module at the end of the pipeline to construct sRGB-to-sRGB pairs for training. As shown in \textbf{Fig.} \ref{fig:raw2rgb}, the output of the model appears with colored stripes in raw domain reconstruction, which is because the network prediction does not take into account the subsequent operations, resulting in slight errors in the RAW domain are amplified in the subsequent processing. Restoration in sRGB also faces challenges, where the color noise is difficult to eliminate. This situation is because the color correction overlaps the information of each channel, which increases the difficulty of restoration. On the contrary, our method processes the input data in the RAW domain and supervises the output in sRGB, aiming at correcting the prediction error accumulated by the cascade pipeline.

\section{Analysis}\label{sec:6}
\subsection{Comparisons with Built-in ISP}\label{sec:6.1}

To demonstrate the significance of deploying the optical degradation correction, we compare our results with the built-in ISP. As shown in \textbf{Fig.} \ref{fig:isp comparisons}, the same JPEG compression algorithm is applied for the sake of a fair comparison. Due to the additional sharpening, the built-in ISP realizes similar results to ours in the center (\textbf{Fig.} \ref{fig:isp comparisons}-1). Yet this globally consistent operation fails to infer realistic features under severe degradation. Therefore, the advantage of our model is evident when it comes to the edge of photographs. The proposed pipeline adaptively restore the high-frequency details of leaves (\textbf{Fig.} \ref{fig:isp comparisons}-2) and cords (marked with the red arrow in \textbf{Fig.} \ref{fig:isp comparisons}-3) at the edge of FoV, which are smoothed out by the built-in ISP. Therefore, our dynamic model is well compatible with the existing ISP system and has the potential to correct the prediction error accumulated by the cascade pipeline. Moreover, our method endows post-processing with the ability to perform adaptive restoration according to the spatial information and the image feature.


\subsection{Comparisons with deeplearning-based ISP}\label{sec:6.2}
\begin{figure}[ht]
    \centering
    \includegraphics[width=\linewidth]{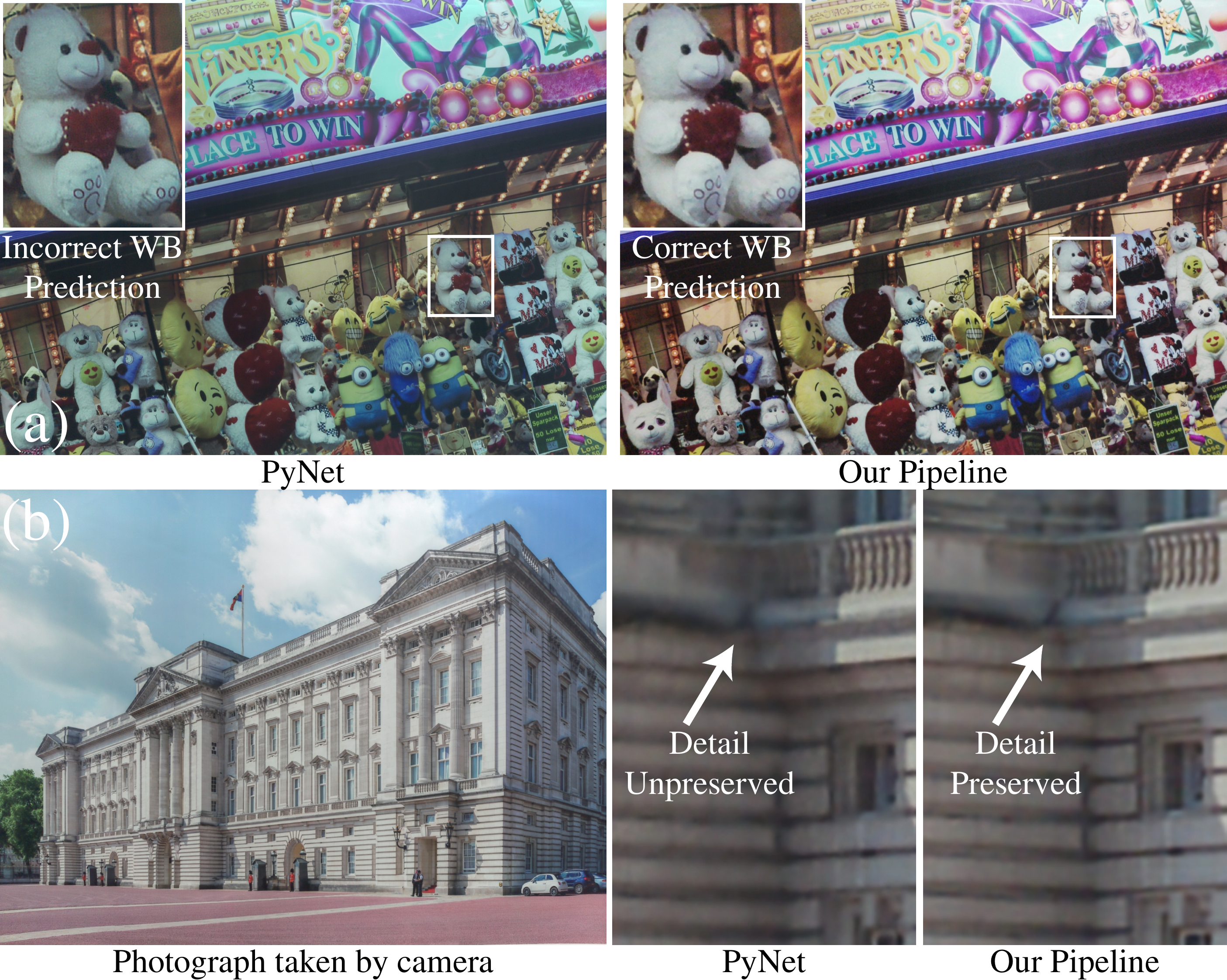}
    \caption{\textbf{Comparisons with the deep-learning ISP}. (a) the failure case of the deep-learning ISP in white balance prediction. (b) evaluation on the ability of detail preservation.}
    \label{fig:pynet comparisons}
\end{figure}

We note that many advanced postprocessing methods attempt to replace the entire ISP system with a deep learning model, called deep ISP. These methods rely on one complex architecture to perform super-resolution (equivalent to demosaic), brightness and color adjustment (equivalent to white balance and color correction), denoising, reconstruction, \textit{etc.} on the captured Bayer image, output the pixel-by-pixel prediction. Due to the need to solve the altogether problems, such deep learning models are engaged with enormous parameters and are rough to implement into mobile terminals. Based on the proposed image formation simulation, we can also synthesize the data pairs from the degraded Bayer raw image to the ideal sRGB ground truth. Therefore, we train the advanced deep ISP model with the synthetic Bayer-to-sRGB data and evaluate the model on real-captured images. The comparisons of the deep ISP model and the proposed method are shown in \textbf{Fig.} \ref{fig:pynet comparisons}. One can see in \textbf{Fig.} \ref{fig:pynet comparisons}a, the PyNet cannot always predict the correct white balance gain when the scene is complicated, resulting in a bluish color of the white bear (magnified in the upper-left corner). Therefore, it is difficult for deep learning models to predict per-channel gain only from input features, let alone some metameric scenes. In view of this, we use the auxiliary white balance from the traditional ISP system to obtain the correct imaging result. We compare the ability of detail resilience in \textbf{Fig.} \ref{fig:pynet comparisons}b when the deep ISP model predicts accurate gain and color correction. As shown, the deep ISP cannot reconstruct detailed textures. This situation is mainly due to the deep ISP model integrating multiple objectives into one model and using the same loss function to guide the restoration. The brightness and color gap between input and ground truth has a crucial influence on the fidelity loss, which will lead the model to eliminate this mismatch in the first place while ignoring the details of the image. In contrast, we separate each task and implement a dynamic model which corrects the postprocessing errors accumulated by the cascaded pipeline. Therefore, our method can accurately restore the color and brightness as well as the detailed information of the scene.

\begin{figure}[t]
    \centering
    \includegraphics[width=\linewidth]{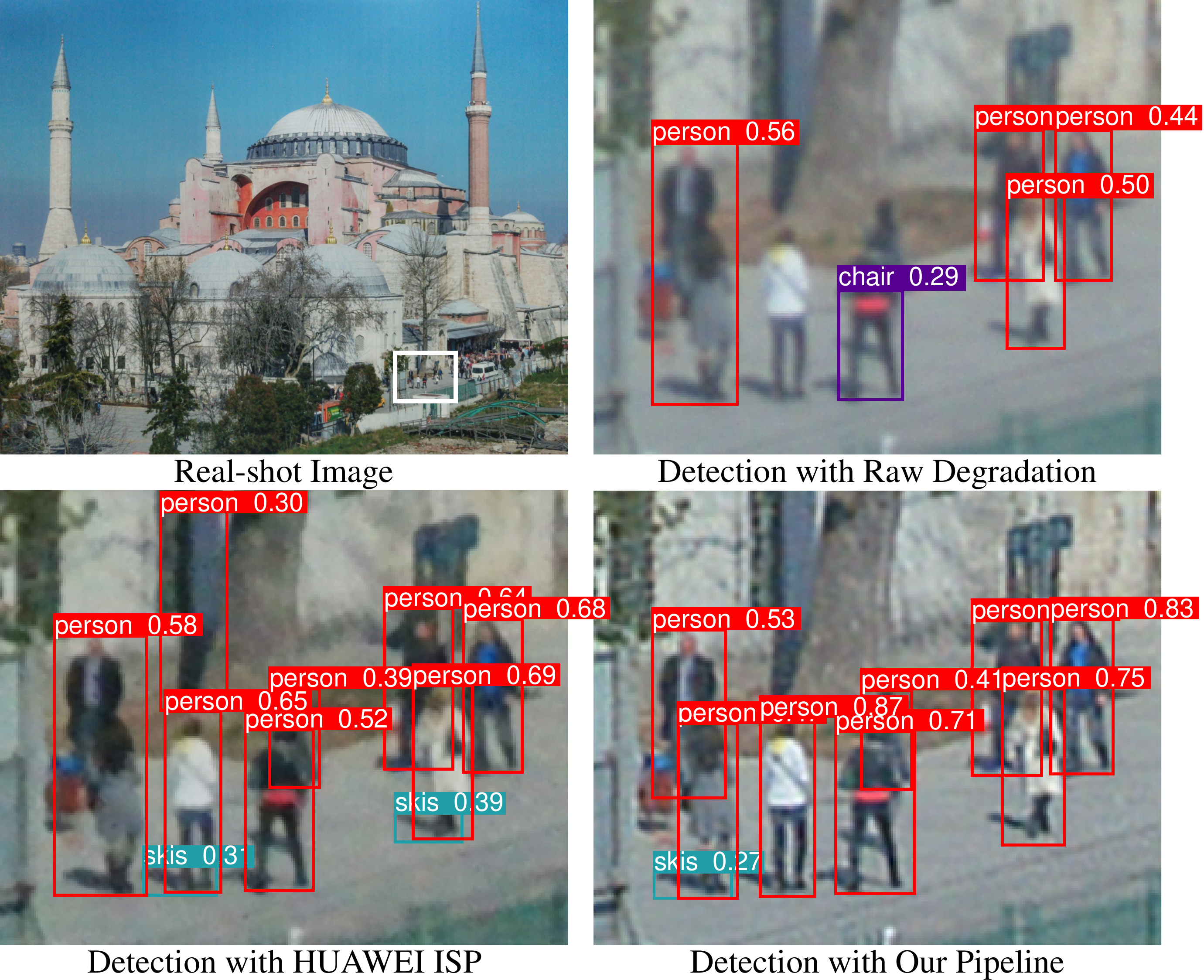}
    \caption{\textbf{Object detection with different post-processing}. The proposed method frees the SOTA detecting algorithms from fine-tuning for a specific camera in the implementation.}
    \label{fig:object detection}
\end{figure}
\begin{figure}[t]
    \centering
    \includegraphics[width=\linewidth]{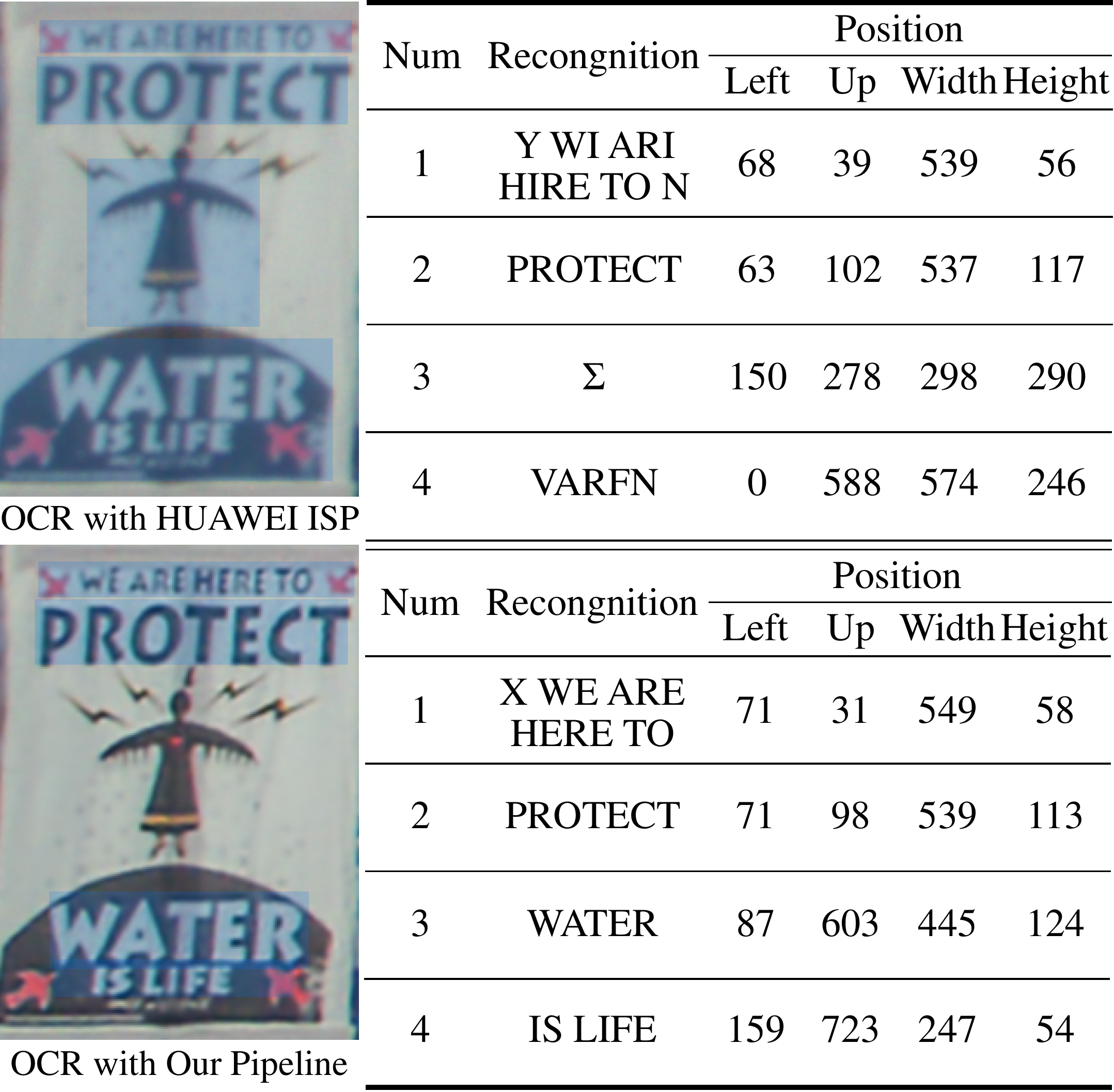}
    \caption{\textbf{OCR with different post-processing}. We help the advanced OCR framework for more accurate recognition and character position.}
    \label{fig:ocr}
\end{figure}

\subsection{Application to Downstream Vision Tasks}\label{sec:6.3}

The advantages of the proposed method, when applied to photographic postprocessing, have been illustrated before. We show that our pipeline could also be helpful for downstream computer vision applications, \textit{e.g.}, object detection, and optical character recognition (OCR). For the evaluation of these tasks, we follow the same pipeline as \textbf{Fig.} \ref{fig:network} to process the captured photographs and conduct off-the-shelf vision algorithms on the processed sRGB images. For comparison, we also show the results on the built-in ISP.

State-of-the-art methods are applied to the daily photos for evaluation, including YOLOv5 \cite{glenn_jocher_2020_4154370} for object detection and PaddleOCR \cite{du2021pp} for OCR. In \textbf{Fig.} \ref{fig:object detection} and \textbf{Fig.} \ref{fig:ocr}, we show the comparisons of visual and numerical results. As shown in \textbf{Fig.} \ref{fig:object detection}, the detection with our pipeline accurately locates all people and gives higher confidence. But the detection with built-in ISP predicts the wrong location to some extent. When applied to OCR, PaddleOCR precisely determine the text regions and recognize text on our result, yet various errors occur in the built-in ISP. In conclusion, our algorithm can directly improve the performance of downstream vision applications, eliminating the need to fine-tune the algorithm for a specific camera.

\section{Conclusion}\label{sec:7}
We presented a perturbed camera model based on the image formation process. And an optimization framework was present to construct a virtual proxy camera from actually measured indicators, whose imaging results are close to the actual manufactured samples. With the proxy cameras from multiple machining samples, we synthesized the data pairs with complex degenerate distributions, aiming at encoding the optical aberrations and the random bias introduced during processing. Drawing from the dynamic convolution, we applied a dynamic model to self-adaptively cope with manufactured cameras, where multiple samples of two typical devices are evaluated to illustrate the benefits of the proposed pipeline. By training only with synthetic data, we demonstrated that our method successfully handles the system of complex machining deviations, achieving perfect restoration that outperforms the high-end DSLR camera.

Our work bridges the gap between optical design, system manufacturing, and postprocessing pipeline. It is convenient to deploy the proposed model in the mass production of arbitrary imaging devices. This work has been applied to some mobile terminals to realize significantly improved imaging. Nevertheless, some challenges remain unsolved to deploy our technique in production. First, for the specific system parameters and manufacturing methods (grinding for DSLR lenses and injection molding for cellphones), the introduced biases are complicated, rendering manual adjustment of damping factors in optimization. Not only that, many existing mobile terminals do not equip with the optimization for operations other than convolution, \textit{e.g.}, attention, which leads to low efficiency in mobile cameras. In conclusion, it is imperative to deploy an optical degradation correction module to the ISP system. As a bridge connecting the hardware system and the algorithm, it is of great help to improve the imaging quality and facilitate the downstream computer vision applications.

\ifCLASSOPTIONcompsoc
  \section*{Acknowledgments}
\else
  \section*{Acknowledgment}
\fi

We thank Meijuan Bian from the facility platform of optical engineering of Zhejiang University for instrument support. We also thank Huawei for their support.

\ifCLASSOPTIONcaptionsoff
  \newpage
\fi




\bibliographystyle{IEEEtran}
\bibliography{IEEEabrv, main}
\end{document}